\documentclass[10pt,twocolumn,letterpaper]{article}

\pdfoutput=1
\usepackage{cvpr}
\usepackage{times}
\usepackage{epsfig}
\usepackage{graphicx}
\usepackage{amsmath}
\usepackage{amssymb}

\usepackage{txfonts}
\usepackage{multirow}
\usepackage{pifont}
\newcommand{\cmark}{\ding{51}}
\newcommand{\xmark}{\ding{55}}
\usepackage{subcaption}

\usepackage{framed}
\usepackage[]{algorithm2e}
\usepackage[mathcal]{euscript}
\usepackage[utf8]{inputenc}

\usepackage[colorlinks]{hyperref}

\cvprfinalcopy

\def\cvprPaperID{1919}

\begin{document}

\title{On Vocabulary Reliance in Scene Text Recognition}

\author{Zhaoyi Wan\textsuperscript{\rm 1}\footnotemark[\value{footnote}]\thanks{Authors contribute equally},
Jielei Zhang\textsuperscript{\rm 1}\footnotemark[\value{footnote}],
Liang Zhang\textsuperscript{\rm 2},
Jiebo Luo\textsuperscript{\rm 3},$^\dagger$,
Cong Yao\textsuperscript{\rm 1}\thanks{Corresponding author}\\
\textsuperscript{\rm 1}Megvii,
\textsuperscript{\rm 2}China University of Mining and Technology,
\textsuperscript{\rm 3}University of Rochester\\
i@wanzy.me, \{yctmzjl,yaocong2010\}@gmail.com
, zhangliang04@hotmail.com, jluo@cs.rochester.edu}

\newcommand{\circleone}{\ding{192}\space}
\newcommand{\circletwo}{\ding{193}\space}
\newcommand{\circlethree}{\ding{194}\space}
\newcommand{\circlefour}{\ding{195}\space}
\newcommand{\circlefive}{\ding{196}\space}
\newcommand{\circlesix}{\ding{197}\space}
\newcommand{\circleseven}{\ding{198}\space}
\newcommand{\circleeight}{\ding{199}\space}
\newcommand{\circlenine}{\ding{200}\space}
\newcommand{\circleten}{\ding{201}\space}

\maketitle
\thispagestyle{empty}
\pagestyle{empty}

\begin{abstract}
The pursuit of high performance on public benchmarks has been the driving force for research in scene text recognition, and notable progress has been achieved. However, a close investigation reveals a startling fact that the state-of-the-art methods perform well on images with words within vocabulary but generalize poorly to images with words outside vocabulary. We call this phenomenon ``vocabulary reliance''. In this paper, we establish an analytical framework to conduct an in-depth study on the problem of vocabulary reliance in scene text recognition. Key findings include: (1) Vocabulary reliance is ubiquitous, i.e., all existing algorithms more or less exhibit such characteristic; (2) Attention-based decoders prove weak in generalizing to words outside vocabulary and segmentation-based decoders perform well in utilizing visual features; (3) Context modeling is highly coupled with the prediction layers. These findings provide new insights and can benefit future research in scene text recognition. Furthermore, we propose a simple yet effective mutual learning strategy to allow models of two families (attention-based and segmentation-based) to learn collaboratively. This remedy alleviates the problem of vocabulary reliance and improves the overall scene text recognition performance.
\end{abstract}

\vspace{-3mm}
\section{Introduction} \label{introduction}

As a pivotal task in many visual recognition and comprehension systems~\cite{Yao2014AUF,Long2018TextSnakeAF,khare2019novel,tian2017unified,Liao2019RealtimeST,Liao2019MaskTA}, scene text recognition has been an active research field in computer vision for decades~\cite{long2018scene, ye2014text,Yao2012DetectingTO,Yao2014StrokeletsAL,Shi2016RobustST,Yang2019SymmetryConstrainedRN,Wan2019TextScannerRC}. Recently, the pursuit of high performance on benchmarks has drawn much attention from the community. Driven by deep learning~\cite{Zhou2017EASTAE,shi2017end,cheng2017fan,aster,hu2020gtc} and large volume of synthetic data~\cite{mjsynth,ren2016cnn,zhan2018verisimilar}, the recognition accuracy on standard benchmarks has escalated rapidly. For instance, the accuracy on IIIT-5k~\cite{mishra2012scene} without lexicon has increased from 78.2\%~\cite{shi2017end} to 96.0\%~\cite{hu2020gtc} in a very short period.

\begin{figure}[t]
    \centering
    \vspace{-2mm}
    \includegraphics[width=0.9\linewidth]{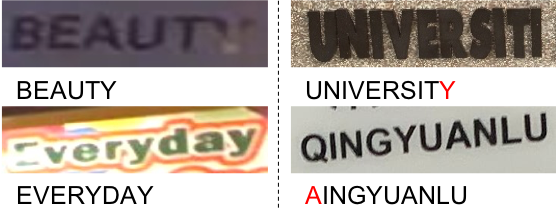}
    \vspace{-3mm}
    \caption{The recurrent memory mechanism in RNN-attention based methods~\cite{aster} is actually a double-edged sword. The positive aspect is that for text images with words in the vocabulary (\textit{Left}), even though image qualities are degraded (blur or partial occlusion), the content can be still correctly recognized. The negative aspect, which is previously neglected, lies in that for text images with words outside the vocabulary (\textit{Right}), mistakes (marked in red) might easily occur.}
    \label{fig:wrong-case}
    \vspace{-5mm}
\end{figure}

However, an important issue has been overlooked for a long time: Even though achieving high accuracy on various benchmarks, state-of-the-art algorithms actually demonstrate obviously higher performance on images with words in the vocabulary\footnote{To be more specific, vocabulary in this work consists of all the words that appear in the training set.} than on those with words outside it. The gap is not caused by the image quality. As shown in Fig.~\ref{fig:wrong-case}, a top-performing text recognizer~\cite{aster} can correctly read the content even for images with poor quality but might make mistakes for images with better quality. The secret lies in the vocabulary: state-of-the-art methods seem inclined to memorize words that have been seen in the training phase. We call this phenomenon ``\textit{vocabulary reliance}''.

To further verify whether \textit{vocabulary reliance} is common in scene text recognition, we reproduce a number of representative methods for scene text recognition, including CRNN~\cite{shi2017end}, FAN~\cite{cheng2017fan}, CA-FCN~\cite{ca-fcn} and ASTER~\cite{aster}. The same backbone network (ResNet-50~\cite{resnet}) and training data (SynthText~\cite{synthtext}) are used for these methods, in order to rule out interference factors. As can be observed from Tab.~\ref{tab:performance-gap}, the performance gaps between test images with words in and outside the vocabulary are significant for all evaluated methods. It reveals that \textbf{\textit{vocabulary reliance is ubiquitous}}.

\begin{table}[t!]
\vspace{-5mm}
    \caption{Accuracy gap between test images with words in and outside vocabulary on IIIT-5k. ``InVoc.'' and ``OutVoc.'' stand for in and outside the vocabulary, respectively.}
    \begin{center}
    \begin{tabular}{|c|c|c|c|c|c|}
    \hline
        Methods  & All   & InVoc.        & OutVoc.       & Gap          \\
    \hline
        CRNN~\cite{shi2017end}      & 86.8  & 91.1        & 68.7         & 22.5          \\
        FAN~\cite{cheng2017fan}     & 89.9       & 93.1         & 75.3        & 17.8          \\
        CA-FCN~\cite{ca-fcn}    & 89.3  & 91.6        & 76.3         & 15.3          \\
        ASTER~\cite{aster}     & 89.2  & 92.9        & 74.6         & 18.4          \\
    \hline
    \end{tabular}
    \end{center}
    \vspace{-5mm}
    \label{tab:performance-gap}
    \vspace{-2mm}
\end{table}

In this paper, we systematically investigate the problem of vocabulary reliance in scene text recognition. An evaluation framework is established, in which training datasets with controlled vocabularies and targeted metrics are devised to assess and compare different module combinations.

Using training data with controlled vocabularies, we are able to inspect the impact of vocabulary on algorithm performance and abilities of different algorithms in learning language prior. Meanwhile, targeted metrics allows for the evaluation of the strengths and weaknesses of different module combinations in a quantitative and precise manner. Through experiments, we obtain a series of valuable observations and findings and accordingly give a few guidelines for choosing module combinations and suggestions for developing scene text recognition algorithms in the future. 

Furthermore, in order to alleviate vocabulary reliance in existing methods, we propose a novel mutual learning strategy, which allows models with different PRED layers, i.e., attention-based decoder and segmentation-based decoder, to complement each other during training. Experimental results demonstrate its effectiveness in improving the accuracy and generalization ability of both attention decoders and segmentation-based methods.

The contributions of this work are as follows:
\vspace{-0.09in}
\begin{itemize}
    \item We raise the problem of vocabulary reliance and propose an analytical framework for investigating it.
    \vspace{-0.08in}
    \item We discovered through experiments the advantages and limitations of current PRED layers. Attention-based decoders generalize poorly from the learned vocabulary but perform well when trained on data with a random corpus. Segmentation-based methods can accurately extract visual features while the CTC- family generally has weaker visual observation ability.
    \vspace{-0.08in}
    \item We found that the effect of CNTX modules, which perform context modeling, is highly coupled with PRED layers. We thus provide guidelines for choosing the CNTX modules according to PRED layers.
    \vspace{-0.08in}
    \item Moreover, we present a simple yet effective mutual learning approach to allow models of different families to optimize collaboratively, which can alleviate the problem of vocabulary reliance.
\end{itemize}

\vspace{-0.1in}
\section{Proposed Analytical Framework} \label{experimental-settings}

In this section, we describe our analytical framework, including data, modules, and metrics, in detail.

\subsection{Test Data}\label{evaluation-data}

To conduct experiments, we adopt various evaluation benchmarks, among which some are commonly used in prior works. We first briefly introduce public test datasets with real-word images, whose details are referred to ~\cite{baek2019wrong}.

\textbf{ICDAR2013 (IC13)}~\cite{ic13} is a dataset of ICDAR 2013 Robust Reading Competition for camera-captured scene text. 
\textbf{ICDAR2015 (IC15)}~\cite{karatzas2015icdar} comes from scene text images collected by Google glasses, where cropped text images are blurred, oriented and with low-resolution. 
\textbf{Street View Text (SVT)}~\cite{wang2011end} is an outdoor street images collection from Google Street View, including noisy, blurry or low-resolution images.
\textbf{SVT Perspective (SVTP)}~\cite{quy2013recognizing} focuses on curved text images. The dataset contains 645 evaluation images, which are severely distorted by non-frontal perspectives.
\textbf{CUTE80 (CT)}~\cite{cute} consists of 80 natural scene images, from which 288 cropped word images are generated for scene text recognition.

Basically, as shown in Fig.~\ref{fig:wrong-case}, the recognition of text images with difficulty in visual features, such as blur, stain, and irregular fonts, relies more on speculation according to the vocabulary. Thus, we group 5 datasets mentioned above into a set $\Omega$. The ground truths of $\Omega$ are collected as our corpus for synthetic training data. Therefore, $\Omega$ and its complement $\Omega^c$ stand for the set of text images in and outside vocabulary, respectively.

Another evaluation dataset, \textbf{IIIT-5k (IIIT)}~\cite{mishra2012scene}, is excluded from corpus collecting, which generally contains regular text and is of clear appearance. We choose IIIT as the stand-along set to conduct the $\Omega^c$ due to its relatively large amount of images and visual clearance. By the collected vocabulary, 1354 images in vocabulary are divided into $\Omega$ and the left 1646 images make $\Omega^c$. They are named as \textbf{IIIT-I} and \textbf{IIIT-O}, respectively.

The size of the datasets and the number of their vocabularies are shown in Tab.~\ref{tab:word-count}. Besides, there are 3172 distinct words in the vocabulary of $\Omega$.

\subsection{Training Data}

Recent works for scene text recognition use synthetic data~\cite{synthtext,mjsynth} for training. \textbf{SynthText (ST)} is a dataset generated by a synthetic engine proposed in \cite{synthtext}, whose background images are extracted from Google Image Search. It contains 80k images, from which researchers cropped about 7 million text instances for training.

As shown in Tab.~\ref{tab:word-count}, ST is generated from a large corpus from Newgroup20~\cite{Newsgroup20} dataset, which has tens of thousands of words in the vocabulary. The large vocabulary of ST obfuscates the impact and cause of vocabulary reliance on such training data.
Therefore we generate new training data for study by constraining the vocabulary.

\begin{figure}[t]
    \vspace{-1mm}
    \centering
    \vspace{-2mm}
    \includegraphics[width=0.9\linewidth]{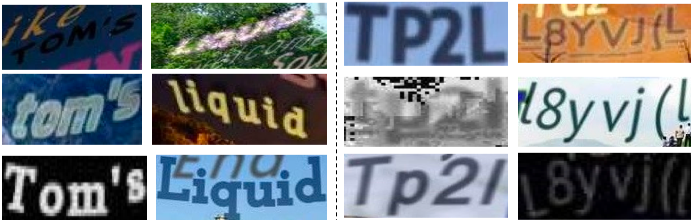}
    \caption{Samples of generated training data. From top to bottom: all uppercase, all lowercase, and the first-letter-capitalized case. The left 2 columns are images picked up from LS, while the right 2 columns are ones from RS.}
    \label{fig:trainingdataset}
    \vspace{-1mm}
\end{figure}

Specifically, as stated in Sec.~\ref{evaluation-data}, our corpus is collected from test datasets. Using the synthetic engine of ST, three datasets with a similar appearance and diverse corpus are conducted for thorough and controlled comparison. Examples are illustrated in Fig.~\ref{fig:trainingdataset}.

\noindent\textbf{LexiconSynth (LS)}
From collected ground truth words, we build the corpus for LS by uniformly sampling from instances. As the vocabulary of $\Omega$ is covered by LS, models trained with LS data acquire the facilitation of vocabulary learning when evaluated on $\Omega$. However, this purified corpus also exacerbates the over-fitting to words in vocabulary. In observation of the performance gap, properties about vocabulary learning of models can be dogged out.

\noindent\textbf{RandomSynth (RS)}
In contrast to LS, the corpus of RS data is generated from characters in a random permutation. The lengths of the pseudowords are of the same distribution with those in LS, but the distribution of character classes is uniform. That is, the accuracy of models trained on RS is achieved without the assistance of vocabulary prior.

\noindent\textbf{MixedSynth (MS)}
An intuitive solution for preventing algorithms from vocabulary reliance is to mix RS data into LS data. In our experiments, MS data is the union of LS and RS. Instances are sampled from RS and LS with ratio $r:(1-r), r\in [0,1]$. The training steps are fixed in all experiments. In comparison with datasets with a large vocabulary, the mixture of RS and LS is more practicable in real-world situations where the vocabulary is seldom completely given in advance.

\begin{table}[t]
    \vspace{-4mm}
    \caption{The number of words and images in training and evaluation data. ``Voc.'' is the vocabulary of datasets. ``Test'' is the vocabulary collected from test images except IIIT.}
    \vspace{-6mm}
    \begin{center}
    \begin{tabular}{|c|c|c|c|c|c|}
    \hline
        \multirow{2}{*}{Dataset} & \multirow{2}{*}{Voc.} & \multicolumn{2}{c|}{Images} & \multicolumn{2}{c|}{Words} \\ \cline{3-6}
                        &           & \small{InVoc.}    & \small{OutVoc.}   &  \small{InVoc.} & \small{OutVoc.} \\

    \hline
         ST   & ST                 & \small{7266715}    &     -    &    76222     &   -  \\
        IC13 & ST                  &  857     & 158      & 549   & 142    \\
        IC15 & ST                  &  1369    & 442      & 669   & 348   \\
        SVT  & ST                  &  530     & 117      & 333   & 94     \\
        SVTP & ST                  &  536     & 109      & 300   &  80    \\
        CT   & ST                  &  218     & 70       & 171   & 63      \\
        IIIT & ST                  &  2429    & 571      & 1277  & 495    \\
    \hline
        IIIT & Test                &  1354   & 1646      & 502   & 1270    \\
    \hline
    \end{tabular}
    \end{center}
    \label{tab:word-count}
    \vspace{-2em}
\end{table}

\noindent\textbf{Synthesis Details}
As the annotation of evaluation datasets serves in different manners on how to treat the case and punctuation of words, we collect the corpus as case-insensitive words without punctuation. During the rendering of LS data, each gathered word generates three instances with different variants: Uppercase, lowercase, and first-letter-capitalized case. Besides, words are inserted with a randomly chosen punctuation by a chance of 10\%.

For the corpus of RS data, the proportion of letters, digits, and punctuation is about 6:3:1. Each word is rendered in the same three cases as LS data. Following the scale of ST, about 7 million cropped images are generated for RS and LS data respectively. When without special statements, the ratio $r$ of MS data is set as $0.5$ empirically.

\subsection{Module Combinations}\label{analysis}

\begin{figure*}[ht]
    \vspace{-5mm}
    \begin{framed}
  
    \begin{subfigure}{\linewidth}
    \centering
        \includegraphics[width=0.75\textwidth]{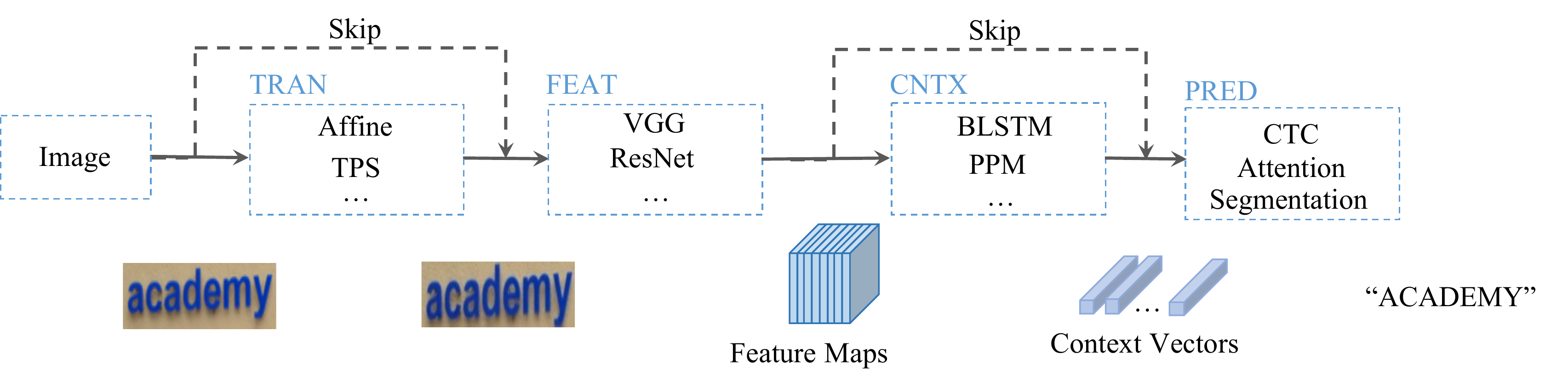}
        \caption{The frameworks of common scene text recognition methods.}\label{fig:pipelines}
    \end{subfigure}
    
    \begin{subfigure}[b]{0.3\linewidth}
    \centering
        \includegraphics[width=0.75\textwidth]{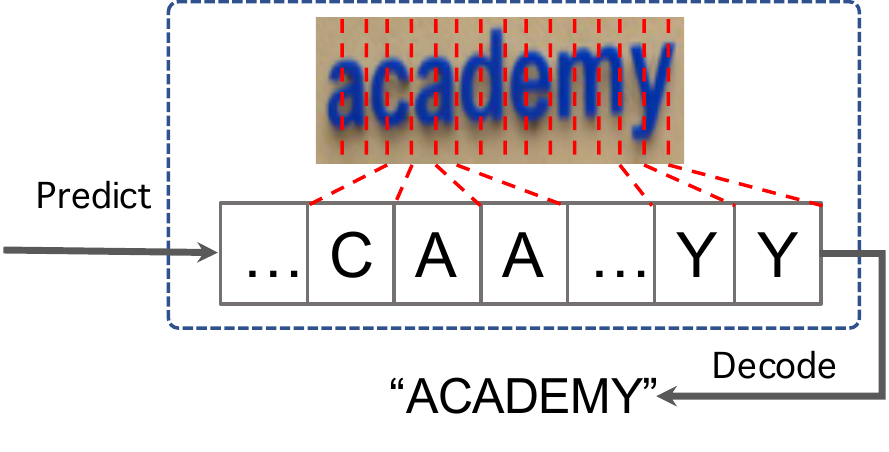}
        \caption{CTC-based decoder.}\label{fig:CTC}
    \end{subfigure}
    \qquad
    \begin{subfigure}[b]{0.3\linewidth}
    \centering
        \includegraphics[width=0.75\textwidth]{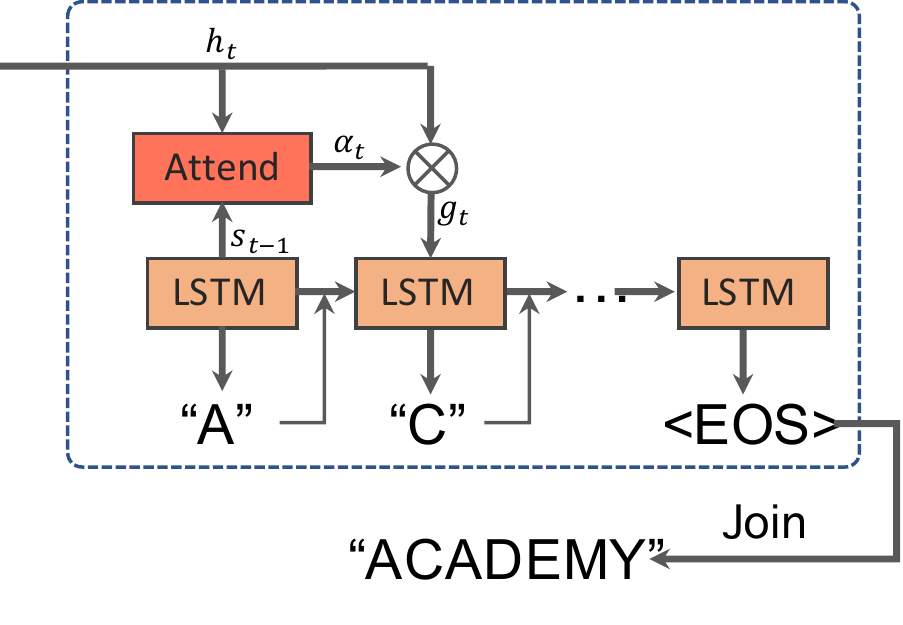}
        \caption{Attention-based decoder.}\label{fig:Attention}
   \end{subfigure}
   \qquad
   \begin{subfigure}[b]{0.27\linewidth}
    \centering
        \includegraphics[width=0.75\textwidth]{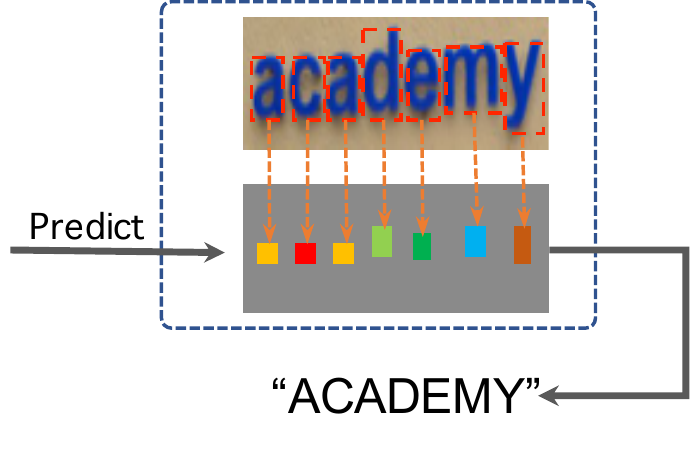}
        \caption{Segmentation-based decoder.}\label{fig:CTC}
    \end{subfigure}
   
    \begin{subfigure}[b]{0.44\linewidth}
    \centering
        \includegraphics[width=0.75\textwidth]{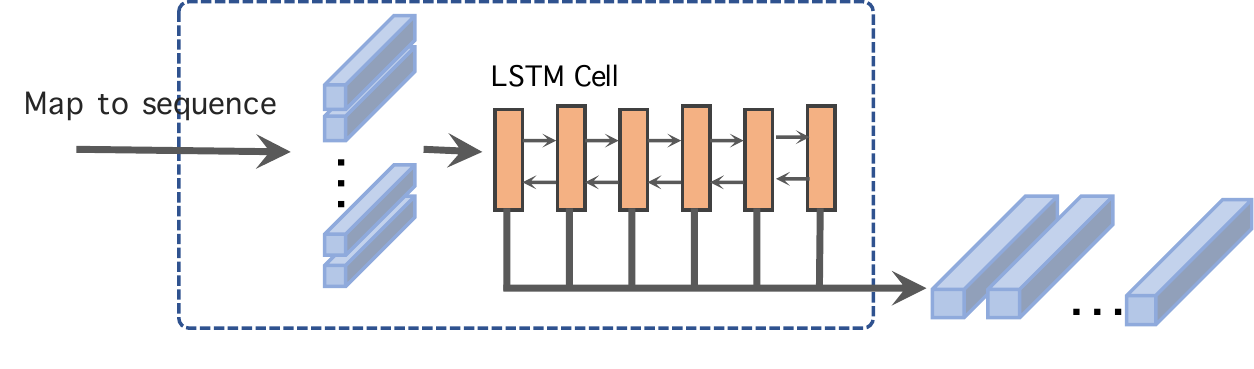}
        \caption{Bi-directional LSTM (BLSTM).}\label{fig:BLSTM}
    \end{subfigure} %
    \qquad
    \begin{subfigure}[b]{0.44\linewidth}
    \centering
        \includegraphics[width=0.75\textwidth]{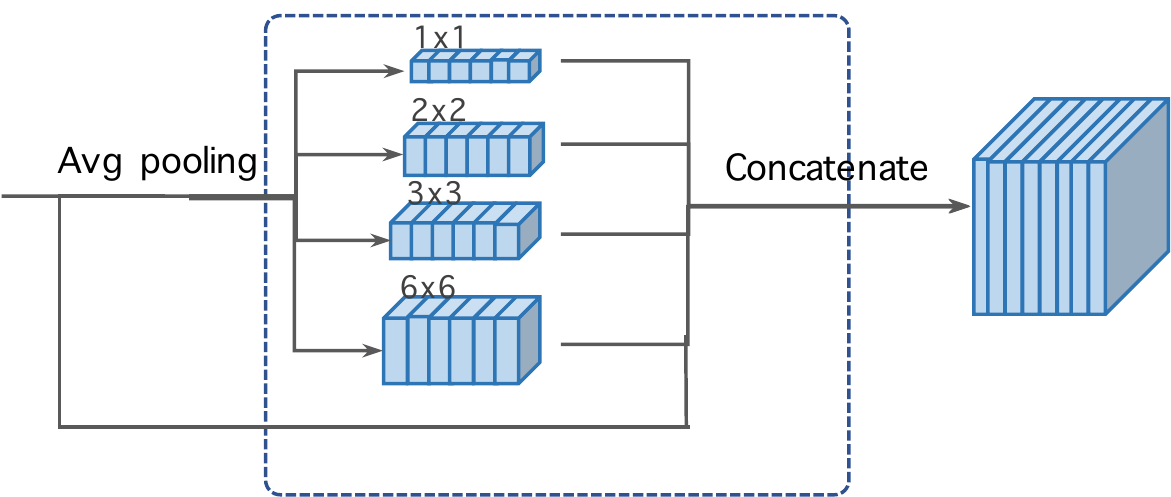}
        \caption{Pyramid pooling module (PPM).}\label{fig:PPM}
   \end{subfigure}
\end{framed}
    \vspace{-5mm}
    \caption{Pipeline and typical modules applied in scene text recognition. ``Skip'' indicates stages which are not requisite thus can be omitted in specific recognition methods.}
    \vspace{-5mm}
    \label{fig:piplelines}
\end{figure*}

\begin{table*}[ht]
    \vspace{-5mm}
    \caption{The raw accuracy of models, which are numbered with circled numbers. ``Gap'' is the accuracy gap between IIIT-I and IIIT-O. ``NGap'' is normalized by recognition accuracy on IIIT.}
    \vspace{-3mm}
    \centering
    \begin{tabular}{|c|c|c|c|ccccccc|c|c|}
        \hline
        \multirow{2}{*}{PRED} & \multirow{2}{*}{CNTX} & \multirow{2}{*}{No.} & \multirow{2}{*}{Data.} & \multicolumn{7}{|c|}{$\subseteq \Omega$}     &  $\subseteq \Omega^c$   &	\multirow{2}{*}{Gap/NGap}   \\ \cline{5-12}
        & & & & AVG & IC13 &	IC15 &	SVT &  SVTP  & CUTE & IIIT-I &  IIIT-O & \\
        \hline
        \multirow{9}{*}{Atten.}    &   
            \multirow{3}{*}{None}   &
            \multirow{3}{*}{\circleone} 
            &   RS      &   68.5	&   82.2	&   55.1	&   71.7    &   57.0    &   54.2   &  83.2 & 73.3 & 9.8/12.6   \\   &    &
            &   MS      &   81.8	&   89.9	&   72.2	&   86.4    &   75.2    &   65.6   &  93.0 & 80.1 & 12.9/15.0   \\   &    &
            &   LS      &   85.7	&   92.7	&   77.4	&   90.5    &   82.3    &   71.5   &  93.7 & 61.0 & 32.7/43.2   \\   \cline{2-13} &
            \multirow{3}{*}{PPM} &    
            \multirow{3}{*}{\circletwo} 
            &   RS      &   70.3	&   84.6	&   57.1	&   74.1    &   58.2    &   55.2   &  84.7 & 77.5 & 7.3/9.0   \\   &    &
            &   MS      &   81.6	&   88.6	&   71.8	&   85.0    &   75.6    &   71.9   &  92.8 & 80.7 & 12.2/14.2   \\   &    &
            &   LS      &   85.5	&   92.1	&   77.0   	&   89.4    &   81.8    &   74.0   &  94.2 & 69.5 & 24.7/30.7   \\   \cline{2-13} &  
            \multirow{3}{*}{BLSTM}   &
            \multirow{3}{*}{\circlethree} 
            &   RS      &   68.6	&   82.4	&   55.4	&   70.9    &   57.0    &   53.5   &  82.9 & 73.8 & 9.4/12.0   \\   &    &
            &   MS      &   82.7	&   89.3	&   74.5	&   86.6    &   77.8    &   67.0   &  92.7 & 81.0 & 11.7/13.6   \\   &    &
            &   LS      &   87.0	&   92.7	&   79.8	&   92.0    &   84.2    &   73.3   &  94.2 & 63.9 & 30.3/39.1   \\   \cline{1-13}
        \multirow{9}{*}{CTC}    &   
             \multirow{3}{*}{None}  &
             \multirow{3}{*}{\circlefour} 
            &   RS      &   64.1	&   80.4	&   47.8	&   66.1    &   49.1    &   55.2   &  81.8 & 71.5 & 10.3/13.5   \\   &    &
            &   MS      &   69.8	&   81.0	&   56.5	&   72.7    &   57.6    &   57.6   &  86.7 & 74.3 & 12.4/15.5   \\   &    &
            &   LS      &   77.8	&   87.0	&   65.8	&   81.9    &   68.8    &   66.0   &  91.6 & 73.6 & 18.0/22.0   \\   \cline{2-13} &
            \multirow{3}{*}{PPM}    &
            \multirow{3}{*}{\circlefive} 
            &   RS      &   62.5	&   76.5	&   48.0	&   62.8    &   47.2    &   49.0   &  81.6 & 68.0 & 13.6/18.5   \\   &    &
            &   MS      &   75.9	&   86.2	&   64.2	&   79.2    &   64.5    &   62.1   &  90.6 & 77.0 & 13.6/16.3   \\   &    &
            &   LS      &   84.8	&   90.9	&   76.0	&   89.8    &   79.2    &   76.0   &  94.2 & 70.1 & 24.1/29.8   \\   \cline{2-13} &  
            \multirow{3}{*}{BLSTM}  &
            \multirow{3}{*}{\circlesix}
            &   RS      &   66.1	&   81.2	&   52.3   	&   67.9    &   51.9    &   51.4   &  82.4 & 72.6 & 9.8/12.7   \\   &    &
            &   MS      &   74.9	&   85.9	&   62.0	&   77.5    &   64.5    &   62.5   &  90.0 & 78.3 & 11.8/14.1   \\   &    &
            &   LS      &   80.0	&   88.1	&   69.3	&   82.7    &   71.6    &   68.8   &  93.1 & 73.5 & 19.6/23.8   \\   \cline{1-13}
        \multirow{6}{*}{Seg.}    &   
            \multirow{3}{*}{None}   &
            \multirow{3}{*}{\circleseven} 
            &   RS      &   68.9	&   80.4	&   56.1	&   71.6    &   57.9    &   55.2   &  84.2 & 73.3 & 10.9/13.9   \\   &    &
            &   MS      &   76.9	&   85.4	&   65.7	&   81.5    &   66.4    &   64.2   &  91.2 & 80.6 & 10.6/12.4   \\   &    &
            &   LS      &   79.7	&   88.4	&   68.7	&   85.7    &   72.1    &   62.2   &  92.3 & 78.8 & 13.5/15.9   \\   \cline{2-13} &
            \multirow{3}{*}{PPM}    &
            \multirow{3}{*}{\circleeight}
            &   RS      &   69.3	&   82.4	&   56.5	&   70.5    &   56.8    &   59.0   &  84.5 & 74.4 & 10.1/12.8   \\   &    &
            &   MS      &   77.6	&   87.3	&   66.8	&   81.5    &   67.1    &   64.2   &  90.9 & 79.9 & 11.0/13.0   \\   &    &
            &   LS      &   81.6	&   89.3	&   72.3	&   85.8    &   75.2    &   64.6   &  92.9 & 76.8 & 16.1/19.2   \\   \cline{2-13}
            \hline
            \hline
        \multirow{3}{*}{Atten.+Mut.}    &   
            \multirow{3}{*}{None}   &
            \multirow{3}{*}{\circlenine} 
            &   RS      &   70.4	&   82.8	&   57.0	&   72.7    &   58.8    &   56.9   &  86.3 & 75.8 & 10.5/13.1   \\   &    &
            &   MS      &   82.0	&   89.9	&   72.3	&   86.4    &   75.2    &   68.1   &  93.1 & 80.7 & 12.4/14.3   \\   &    &
            &   LS      &   85.8	&   91.9	&   77.2	&   90.8    &   83.1    &   72.7   &  94.5 & 77.6 & 16.9/19.9   \\   \cline{2-13}
            \hline
        \multirow{3}{*}{Seg.+Mut.}    &   
            \multirow{3}{*}{None}   &
            \multirow{3}{*}{\circleten}
            &   RS      &   70.0	&   82.4	&   56.1  	&   70.8    &   57.4    &   59.0   &  84.3 & 74.7 & 10.0/12.1   \\   &    &
            &   MS      &   78.3	&   87.8	&   66.7	&   82.1    &   67.7    &   68.0   &  91.2 & 79.3 & 12.4/14.4   \\   &    &
            &   LS      &   82.3	&   89.4	&   71.3	&   86.4    &   78.6    &   72.5   &  93.6 & 80.0 & 13.6/15.7   \\   \cline{2-13}
            \hline
    \end{tabular}
    \vspace{-2mm}
    \label{tab:module-performance}
    \vspace{-4mm}
\end{table*}

According to~\cite{baek2019wrong}, a typical scene text recognition method can be divided into four stages, \textit{transformation (TRAN)}, \textit{feature extraction (FEAT)}, \textit{context modeling (CNTX)}, and \textit{prediction (PRED)}. The CNTX stage is similar to sequence modeling (Seq.) in \cite{baek2019wrong}. We extend to modeling context as we also take segmentation-based methods into consideration, for the sake of discussing the problem of vocabulary reliance in a broader perspective. The pipeline of scene text recognition is shown in Fig.~\ref{fig:piplelines}. 

In our experiments and analyses, we focus on CNTX and PRED stages, as these two stages are highly relevant to vocabulary reliance. TRAN and FEAT stages are fixed to control variables: No transformation layer is adopted and ResNet50 backbone is used in all combinations. Below, we will introduce three PRED layers and three choices for the CNTX stage.

\noindent\textbf{Prediction Layers}
CTC~\cite{ctc} and attention-based decoders~\cite{aon,yang2019symmetry} are two dominating approaches in the choices of prediction layers. As illustrated in Fig.~\ref{fig:CTC}, CTC aligns the frame-wise predictions into the target string. Frames with the same characters without ``BLANK'', which is introduced to stand for no characters, are removed in final outputs. CTC is widely used in many real-world applications~\cite{li2018toward} and academic researches~\cite{nlpr_attn_ctc,sliding_lstm_ctc2016}, due to its superior inference speed~\cite{baek2019wrong}.

Attention-based (Atten. for short) decoders~\cite{cheng2017fan,aster} are state-of-the-art methods in the field of scene text recognition. A glimpse vector is generalized from the feature sequence, upon which an RNN is adopted to produce attention vectors over the feature sequence and produce character classification each in order~(see Fig.~\ref{fig:Attention}).

Recently, MaskTextSpotter~\cite{mask_textspotter} introduces instance segmentation to localize and classify each character separately and inspires following works~\cite{gao2019reading,ca-fcn,fcn_attn}.
Although segmentation-based (Seg. for short) methods directly extract characters by finding connected components in the segmentation map, the large receptive field of deep convolutional networks might bring vocabulary reliance.

\noindent\textbf{Context Modules}
Bi-directional LSTM (BLSTM)~\cite{lstm} is employed for context modeling on top of feature maps extracted by CNNs in recent works~\cite{aster,lei2018scene}.

As illustrated in Fig.~\ref{fig:BLSTM}, the BLSTM module takes feature sequences as input, which are transformed from feature maps by pooling or convolution with strides. It is a common practice in many scene text recognition methods~\cite{su2014accurate,yang2019symmetry} for context modeling, as the BLSTM scans and maps features in the bi-directional order.

Pyramid pooling module (PPM)~\cite{pspnet} shown in Fig.~\ref{fig:PPM} is another choice for context modeling, which is proved effective on segmentation-based methods~\cite{krishnan2018word}. It utilizes adaptive average pooling to pool feature maps into different square resolutions (1, 3, 4, 6 in our experiments). Pooled features are then resized to the input resolution by bi-linear interpolation and concatenated with original features to gain global context information in different scales. Since segmentation-based methods are incompatible with BLSTM, PPM is a practical module for context modeling. Our experiments also validate its effectiveness in enhancing the vocabulary learning of models.

Besides, the explicit contextual modeling is not requisite for robust text recognition, as deep convolutional networks usually have large receptive fields~\cite{xie2016fully,zhang2019sequence}. Though, in our experiments, context modeling modules do bring diversity in vocabulary learning and reliance.

The raw results are shown in Tab.~\ref{tab:module-performance}, in which module combinations are named with circled numbers.

\subsection{Metrics}\label{evaluation-metrics}

\begin{table}[t]
    \caption{The computation of proposed metrics. Therein, $Acc(\cdot)$ and $Gap(\cdot)$ are defined in Sec.~\ref{evaluation-metrics}.}
    \vspace{-5mm}
    \begin{center}
    \begin{tabular}{|c|c|}
    \hline
        Metrics.   & Computation          \\
    \hline
        $\mathcal{GA}$	      & $Acc(X_{train}, \Omega \cup \Omega^c)$ \\
        $\mathcal{OA}$	      &	$Acc(RS, \Omega \cup \Omega^c)$ \\
        $\mathcal{VA}$	      & $Acc(LS, \Omega)$ \\
        $\mathcal{VG}$	      &	$1 - (Gap(LS) - Gap(RS))$ \\
        $\mathcal{HM}$	      &	$3(\frac{1}{\mathcal{OA}} + \frac{1}{\mathcal{VA}} + \frac{1}{\mathcal{VG}})^{-1}$ \\
    \hline
    \end{tabular}
    \end{center}
    \label{tab:metrics}
    \vspace{-5mm}
\end{table}
Using our re-designed training data, we can evaluate the performance of algorithms on several training data. Several metrics are proposed for benchmarking the properties of models in aspects.

Firstly, we introduce a conventional metric for performance evaluation, \textbf{General Accuracy ($\mathcal{GA}$)}. The current practice for evaluating algorithms of scene text recognition is to evaluate models on public benchmarks with real-world images. We define the recognition accuracy on all test images of the mentioned evaluation datasets as $\mathcal{GA}$, corresponding to the common evaluation in previous works.

In addition to the general metric, we further propose three specific metrics and their harmonic mean to fully reflect particular properties of different methods. For clarity, let's define two functions. $Acc(X_{train}, X_{test})$ is the accuracy of models trained on dataset $X_{train}$ and tested on dataset $X_{test}$. $Gap(\cdot)$ is defined as the performance gap on IIIT-I and IIIT-O with the same training data $X_{train}$:
\begin{equation}
    \begin{split}
    Gap(X_{train}) = &Acc(X_{train}, IIIT\text{-}I)\\
                     & - Acc(X_{train}, IIIT\text{-}O).    
    \end{split}
    \vspace{-1mm}
\end{equation}

\noindent\textbf{Observation Ability ($\mathcal{OA}$)}
Accurate visual feature extraction and recognition is the fundamental ability of scene text recognition methods. We define $\mathcal{OA}$ as how accurately an algorithm recognizes words without any vocabulary given in training data. In the context of our framework, $\mathcal{OA}$ is measured by evaluating models trained on RS data with test images from all benchmarks (7406 images in total). As the recognition accuracy purely comes from the observation of visual features without learning any vocabulary, it indicates the ability of models to utilize visual observation.

\noindent\textbf{Vocabulary Learning Ability ($\mathcal{VA}$)}
As stated in Sec.~\ref{introduction}, it is likely for algorithms to employ learned vocabulary to refine or constrain recognition results of text images. Similar to $\mathcal{OA}$, $\mathcal{VA}$ is suggested for evaluating the recognition accuracy on limited vocabularies. In our experiments, measuring of $\mathcal{VA}$ is to train models with LS data and evaluate the recognition accuracy on all images in $\Omega$. $\mathcal{VA}$ is meaningful for choosing models in text recognition tasks where lexicon is provided in advance.

\noindent\textbf{Vocabulary Generalization ($\mathcal{VG}$)}

Human beings can easily generalize things from what they learnt, which inspires us to evaluate the vocabulary generalization($\mathcal{VG}$) of an algorithm by measuring the performance of models trained with LS data on words out of vocabulary. In fact, we witness the vocabulary generalization of current recognition methods in our experiments. 
To fairly evaluate $\mathcal{VG}$, the influence of image visual feature on the dataset, which brings an intrinsic gap between two image sets, is supposed to be eliminated. Therefore $\mathcal{VG}$ is indicated by 
\begin{equation}
    \mathcal{VG} = 1 - (Gap(LS)-Gap(RS))
\end{equation}
where the score is subtracted from 1 in order to unify the monotonicity.

\noindent\textbf{Harmonic Mean ($\mathcal{HM}$)}
For a overall metric, the harmonic mean of $\mathcal{OA}$,$\mathcal{VA}$, and $\mathcal{VG}$ is adopted as the summary score: 
\begin{equation}
    \vspace{-2mm}
    \mathcal{HM}=3(\frac{1}{\mathcal{OA}} + \frac{1}{\mathcal{VA}} + \frac{1}{\mathcal{VG}})^{-1}.
\end{equation}

$\mathcal{HM}$ can be taken as a standard for general comparison of different models.

Besides, evaluation on random string can be a metric, however, there is no standard benchmark that contains pure random labels with real-world complexity . Thus, it will not be discussed in this paper.

\begin{table}[t]
    
    \caption{Metrics of models. The circled number corresponds to different combination of different module. No. is referred to Tab.~\ref{tab:module-performance}.}
    \vspace{-6mm}
    \begin{center}
    \begin{tabular}{|c|c|c|ccc|c|}
    \hline
        No. & PRED &   
        $\mathcal{GA}$ &    $\mathcal{VA}$  &   $\mathcal{VG}$  &   $\mathcal{OA}$      &     $\mathcal{HM}$  \\
    \hline
        \circleone  &   Atten. &
        81.0          & 85.7                & 77.1              & 69.6                  & 76.9          \\
        \circletwo  &   Atten. &
        81.3          & 85.5                & 82.6              & \textbf{71.9}         & 79.5          \\
        \circlethree  &   Atten. &
        \textbf{83.1} & \textbf{87.0}       & 79.1              & 69.8                  & 78.0          \\
        \hline
        \circlefour  &   CTC   &
        75.8          & 77.8                & 92.4              & 65.8                  & 77.1          \\
        \circlefive  &   CTC   &  
        80.1          & 84.8                & 89.5              & 63.5                  & 77.5          \\
        \circlesix  &   CTC   &
        78.4          & 79.9                & 90.2              & 67.6                  & 78.1          \\
        \hline
        \circleseven  &   Seg. & 
        80.8          & 79.7                & \textbf{97.3}      & 69.9                 & 80.8           \\
        \circleeight  &   Seg. &
        81.3          & 81.6                & 94.0              & 70.5                  & \textbf{80.9}         \\
    \hline
    \end{tabular}
    \end{center}
    \vspace{-2em}
    \label{tab:model-metrics}
\end{table}

\section{Comparisons and Analyses}
Using our proposed framework in Sec.~\ref{experimental-settings}, we provide comparisons and analyses on various module combinations. Metrics of models are shown in Fig~\ref{tab:model-metrics}. Based on the specific evaluation, we assess and analyze module combinations in different aspects.

\subsection{Effect of Training Data}

\begin{figure}[t]
    \centering
    \includegraphics[width=0.7\linewidth]{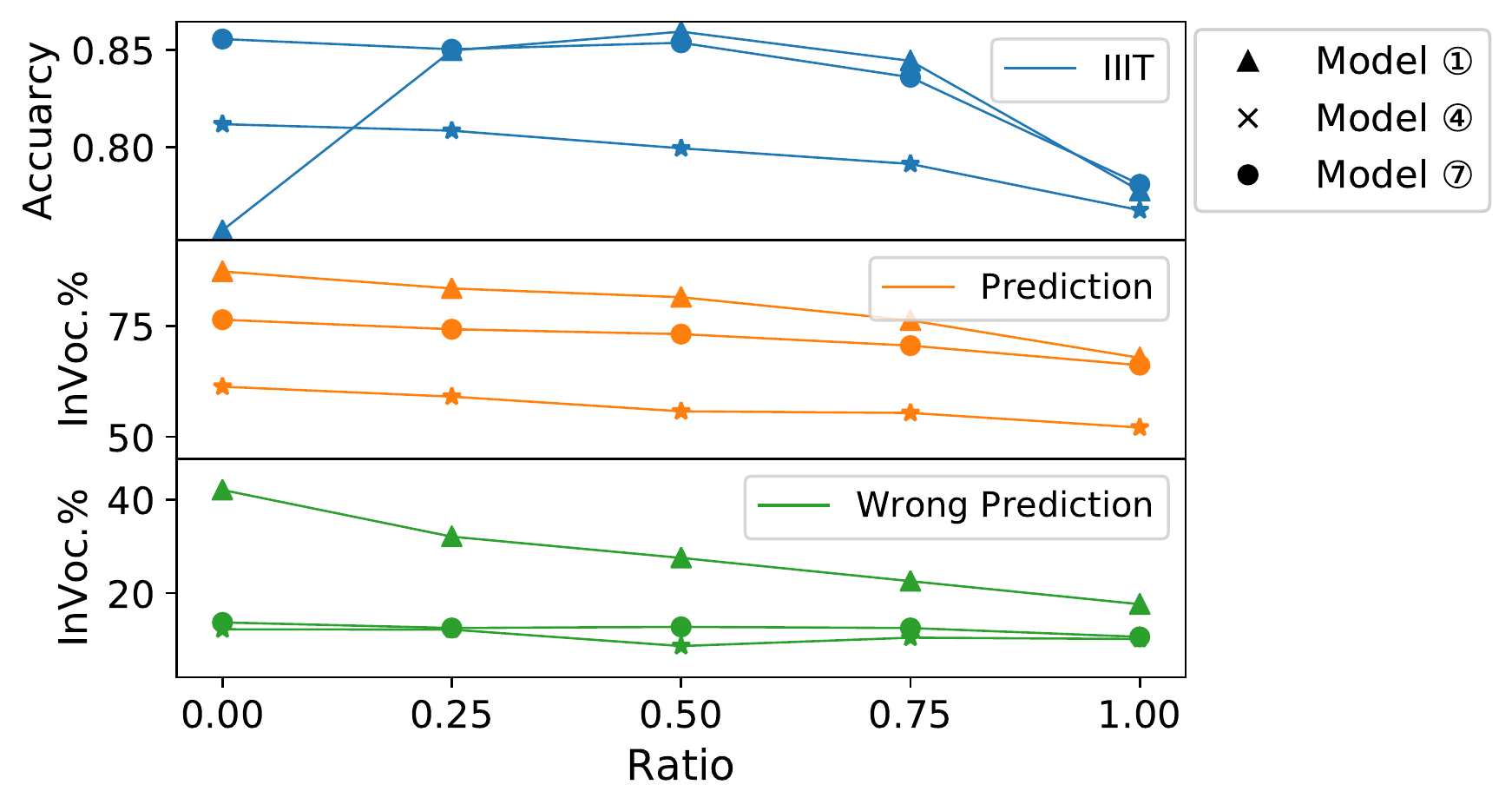}
    
    \caption{Probability of model \circleone,  \circlefour and \circleseven on making prediction inside vocabulary. ``Ratio'' is the ratio of RS in MS data.}
    \label{fig:prediction-in-vocabulary}
    \vspace{-5mm}
\end{figure}

Fundamentally, we should first validate the effectiveness of the proposed dataset and explore the relevance of vocabulary reliance on training data. Experiments are conducted by gradually adjusting the ratio $r$ in MS data from $0$ to $1$. Three models, \circleone, \circlefour and \circleseven in Tab.~\ref{tab:module-performance}, are adopted for comparison. Besides the recognition accuracy on IIIT, we observe the probability of predicted words falling into the vocabulary, as shown in Fig.~\ref{fig:prediction-in-vocabulary}.

With RS data mixed into the LS data, recognition accuracy on IIIT is improved as models trained with the mixed data are less prone to be misled by vocabulary reliance.
Especially for model \circleone, the recognition accuracy on IIIT increases from 77.8\% to 84.4\%,  benefiting from the mixed RS data with a ratio of 25\%.

The improvement in accuracy ceases when $r$ reaches around 0.5. On one hand, the reduction of the probability to produce word prediction in vocabulary proves it effective to countervail vocabulary reliance with RS data. On the other hand, it requires a sufficient ratio of LS data to learn vocabulary from training data.

\subsection{Comparison of Prediction Layers}\label{comparison-prediction-layers}

\begin{figure}[ht]
    \begin{subfigure}{0.453\linewidth}
    \centering
        \includegraphics[width=\linewidth]{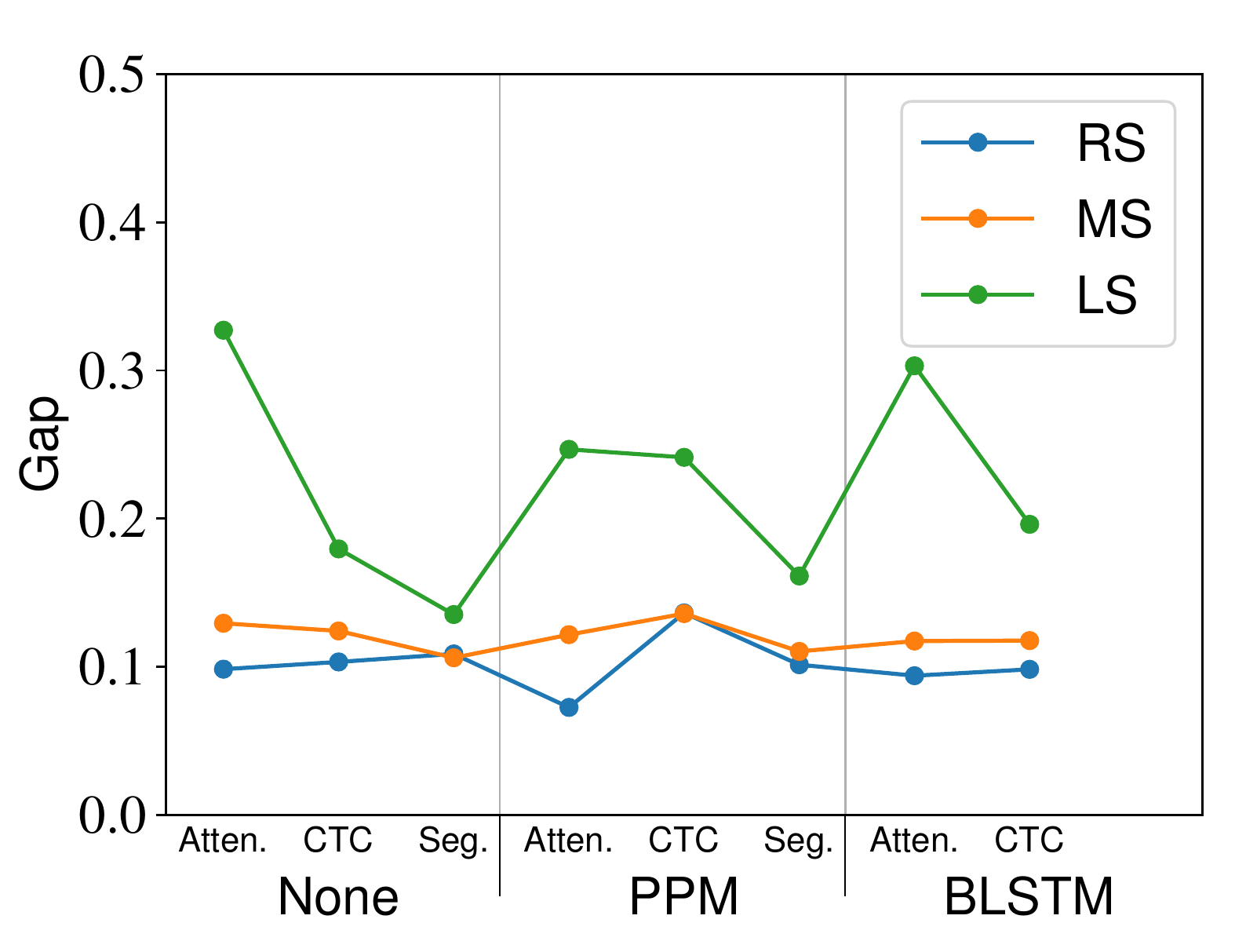}
        \caption{}\label{fig:overall-line}
    \end{subfigure} %
    \qquad
    \begin{subfigure}{0.453\linewidth}
    \centering
        \includegraphics[width=\linewidth]{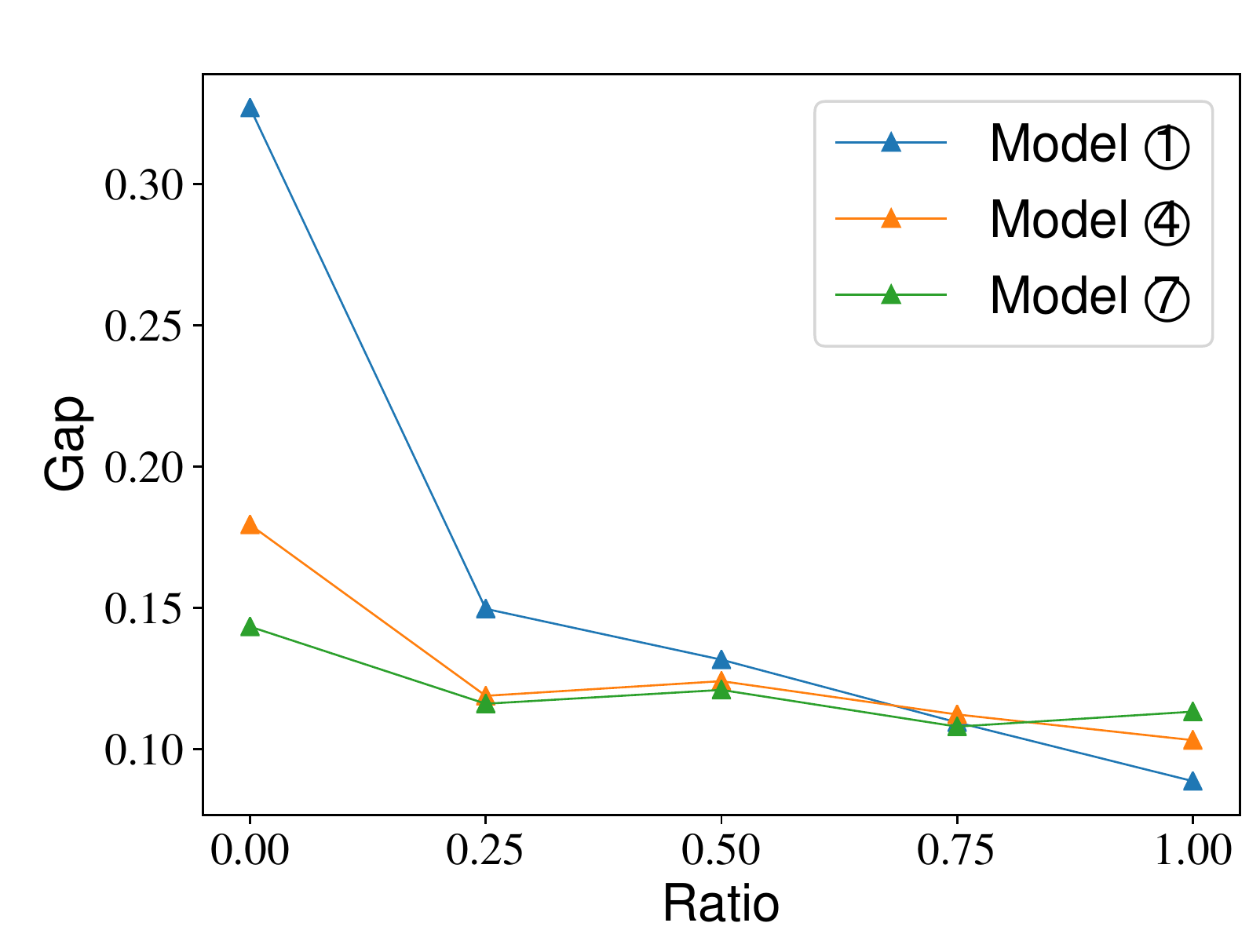}
        \caption{}\label{fig:line_ratio_gap}
   \end{subfigure}
   \vspace{-3mm}
   \caption{The accuracy gap between IIIT-I and IIIT-O. (a) Performance gap on IIIT-I and IIIT-O of module combinations. (b) The gap changes with adjusted ratio of RS data.}
   \vspace{-2mm}
\end{figure}

From Fig.~\ref{fig:overall-line}, we perceive the consistent performance gap between models trained with RS, MS, and LS data, despite PRED layers nor CNTX modules. It shows that all the combinations suffer from the problem of vocabulary reliance, but the severity differs.

\begin{figure}[t]
    \centering
    \begin{subfigure}{0.4\linewidth}
    \centering
        \includegraphics[width=\linewidth]{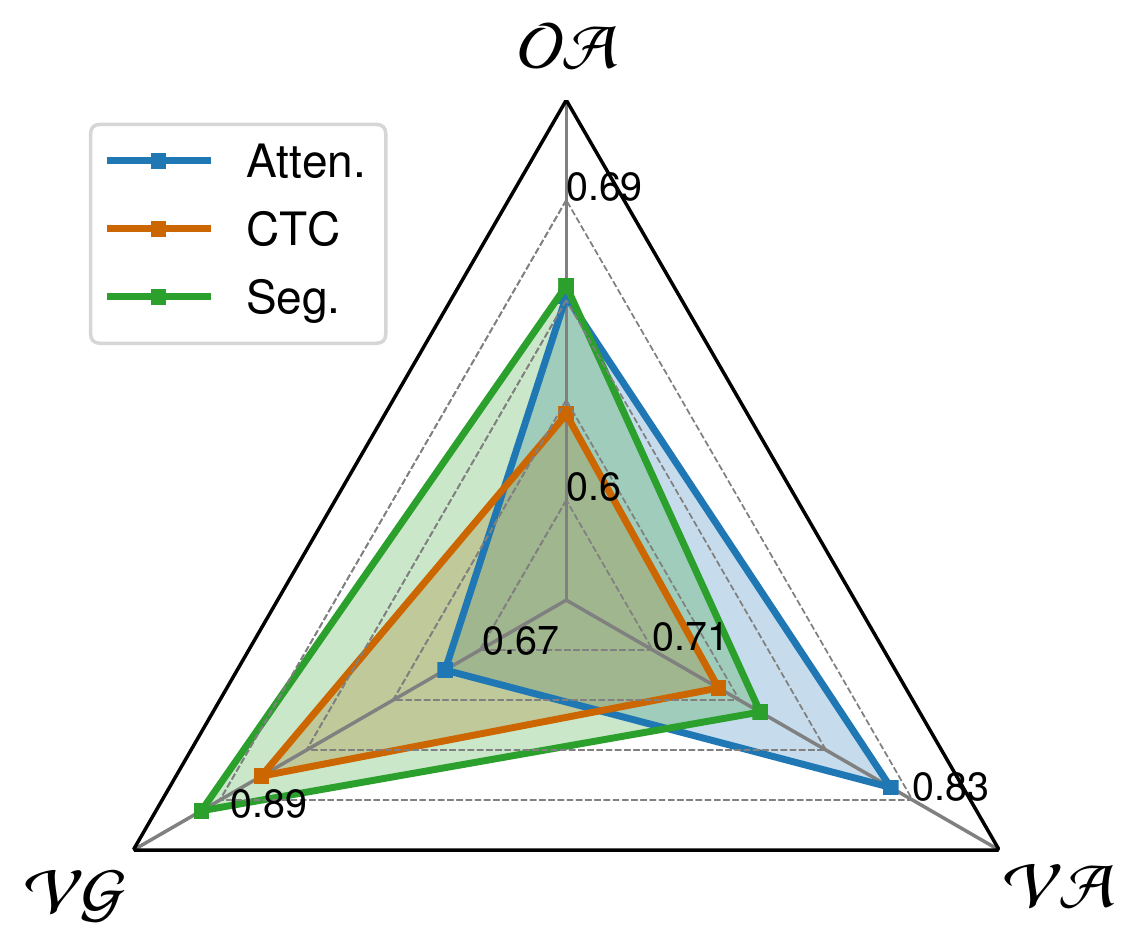}
        \caption{}\label{fig:prediction-layers}
    \end{subfigure} %
    \qquad
    \begin{subfigure}{0.4\linewidth}
    \centering
        \includegraphics[width=\linewidth]{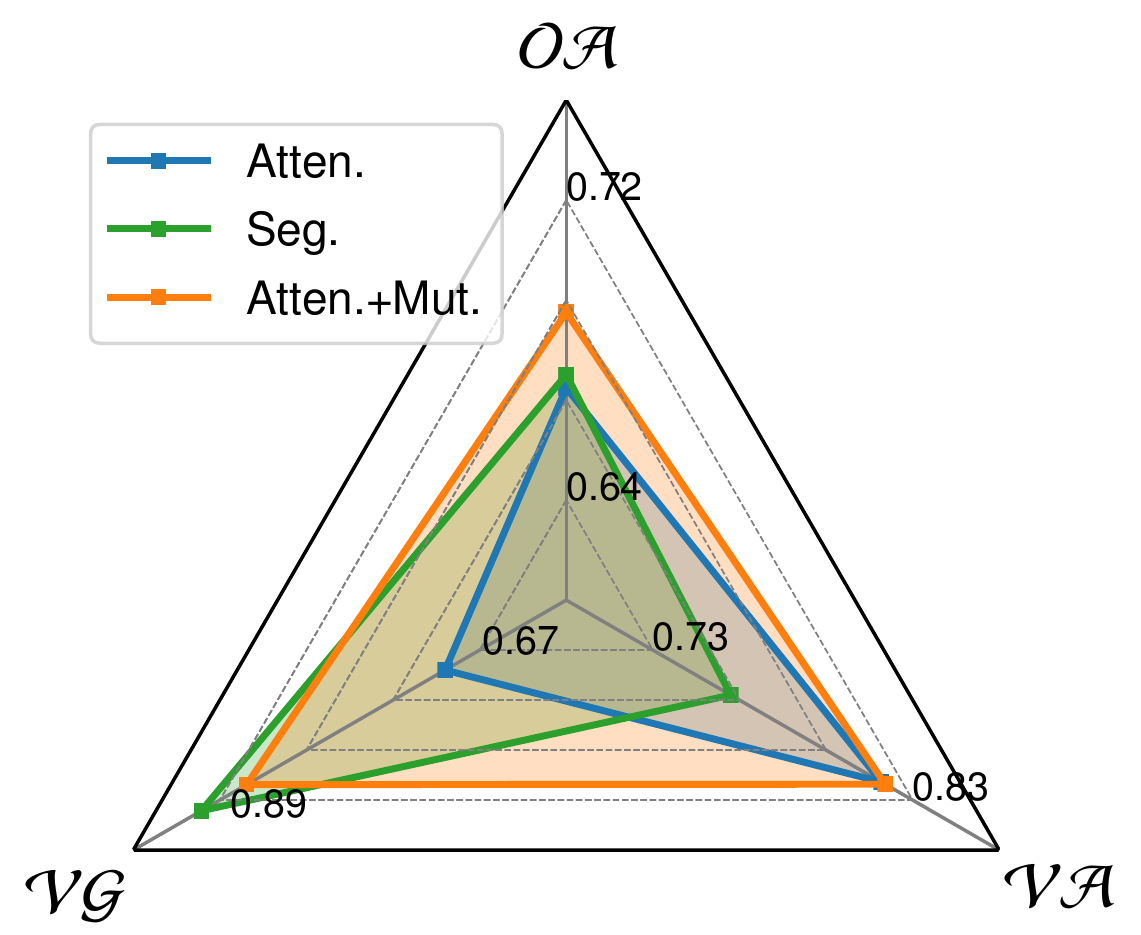}
        \caption{}\label{fig:prediction-mutual}
   \end{subfigure}
   \vspace{-3mm}
   \caption{Performance of PRED layers on our metrics. All models are built without CNTX module. (a) The comparison of PRED layers. (b) $\mathcal{OA}$ and $\mathcal{VA}$ improvement of mutual learning.}
    \vspace{-6mm}
\end{figure}

Moreover, we illustrate the performance gap on IIIT of model \circleone, \circlefour and \circleseven trained with different training data. The models are built without CNTX modules, using the Atten., CTC, and Seg. PRED layers, respectively. The attention-based decoder starts with the highest gap on the point where $r=0$ (LS data), as shown in Fig.~\ref{fig:line_ratio_gap}. With more RS data mixed into the training set, the gap of attention-based decoder decreases. The trend verifies the advantage of attention-based decoders on vocabulary learning and inferiority on vocabulary reliance.

In addition to vocabulary reliance, a thorough comparison of our proposed metrics of the PRED layers is illustrated in Fig.~\ref{fig:prediction-layers}. The performance of CTC is generally covered by the other two prediction layers, on metrics including both accuracy and generalization. Attention-based and segmentation-based decoders gain advantages in $\mathcal{VA}$ and $\mathcal{VG}$ respectively. They also perform similarly well in $\mathcal{OA}$, indicating the ability to accurate recognition according to visual features only.

\subsection{Comparison of Context Modules}

\begin{figure}[ht]
    \begin{center}
    \includegraphics[width=0.6\linewidth]{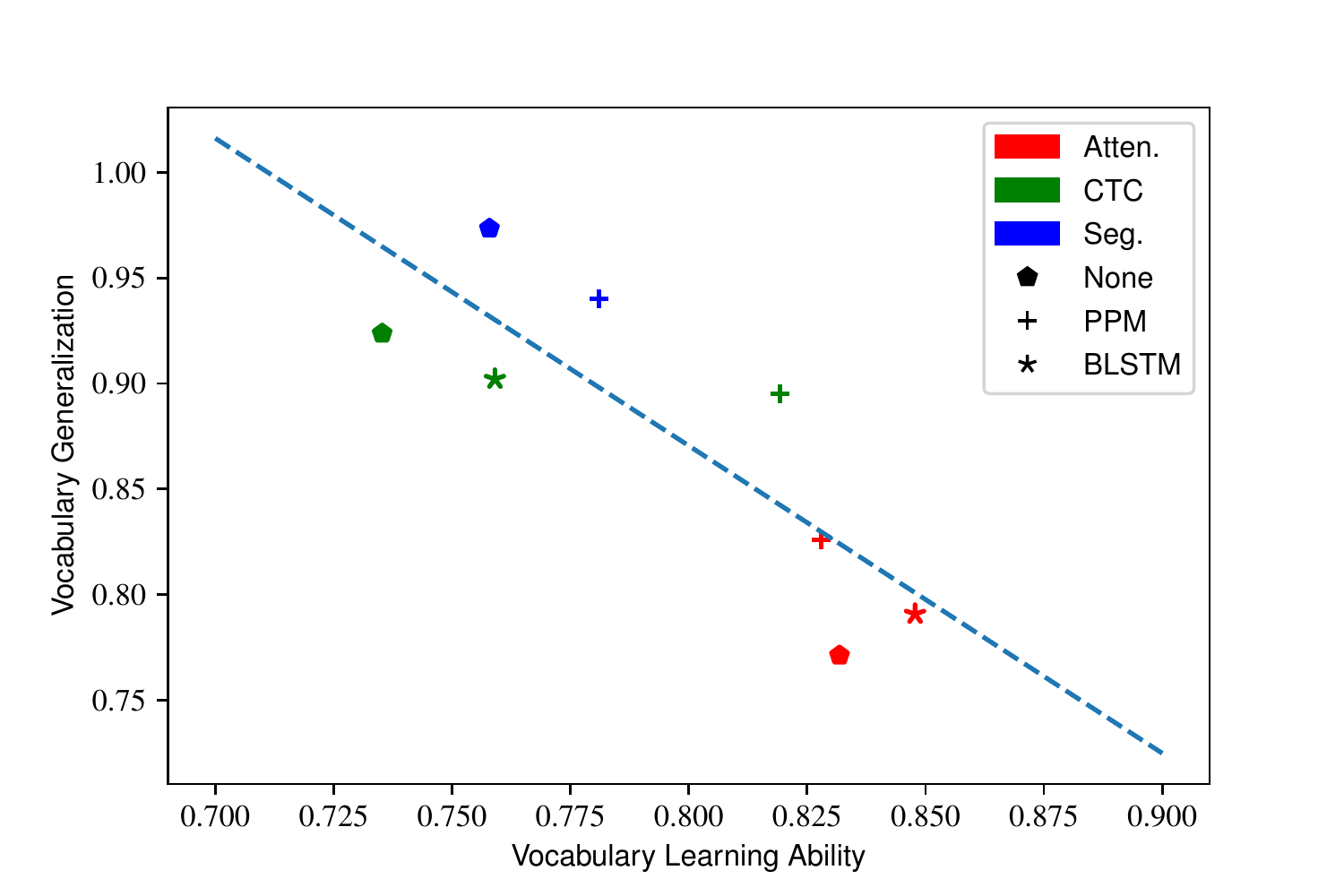}
    \end{center}
    \vspace{-3mm}
    \caption{$\mathcal{VA}$ and corresponding $\mathcal{VG}$ of module combinations.}
    \label{fig:VA-VG}
    \vspace{-6mm}
\end{figure}

Basically, the adoption of context modules improves the vocabulary learning of models, as validated by the $\mathcal{VA}$ of module combinations. For example, PPM, which is not widely used in prior scene text recognition methods, brings boost on $\mathcal{VA}$ in combination with PRED layers: 3.9\% for Seg. and 10.5\% for CTC. On the other hand, as shown in Fig.~\ref{fig:VA-VG}, the strength in $\mathcal{VA}$ usually carries a decrease in $\mathcal{VG}$.

\begin{figure}[t!]
    \begin{subfigure}{0.4\linewidth}
    \centering
        \includegraphics[width=\textwidth]{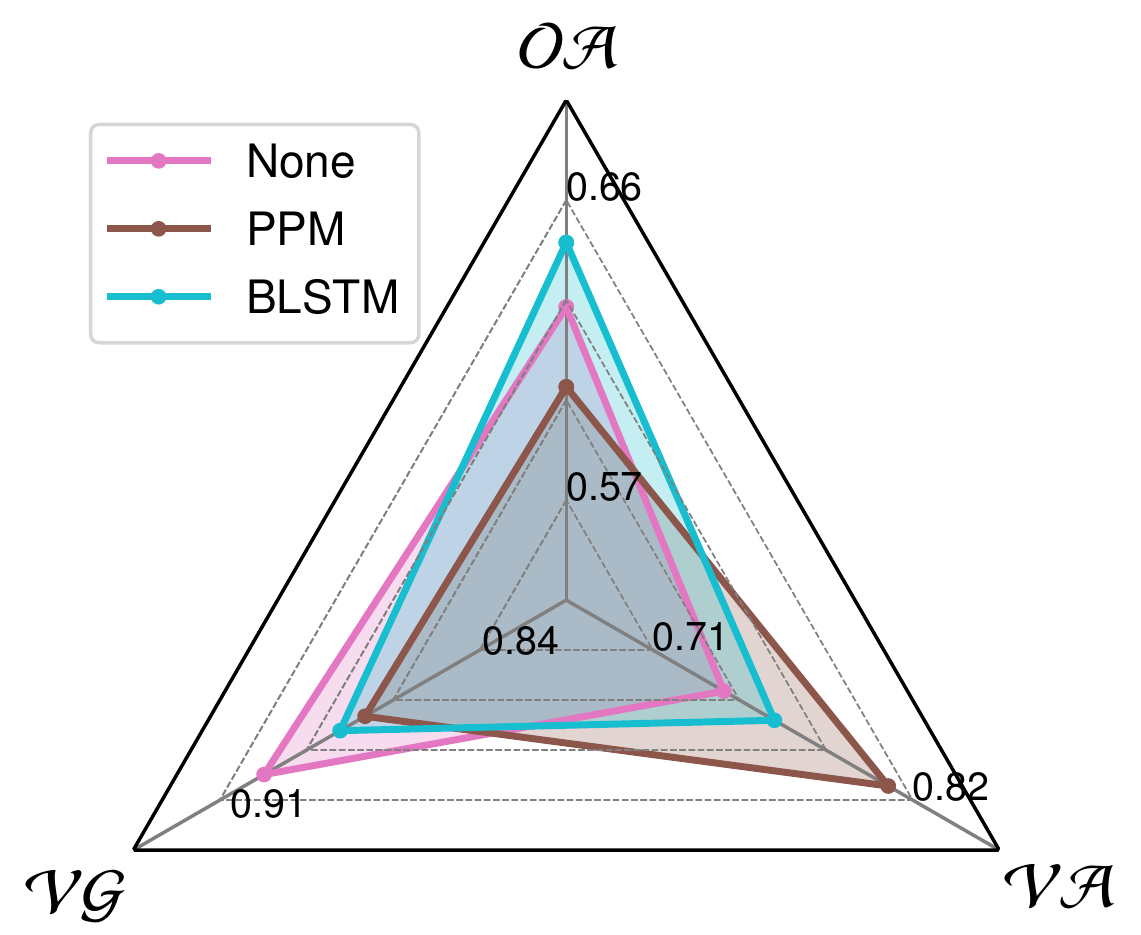}
        \caption{CTC PRED}\label{fig:CTC-pred}
    \end{subfigure}
    \qquad
    \begin{subfigure}{0.4\linewidth}
    \centering
        \includegraphics[width=\textwidth]{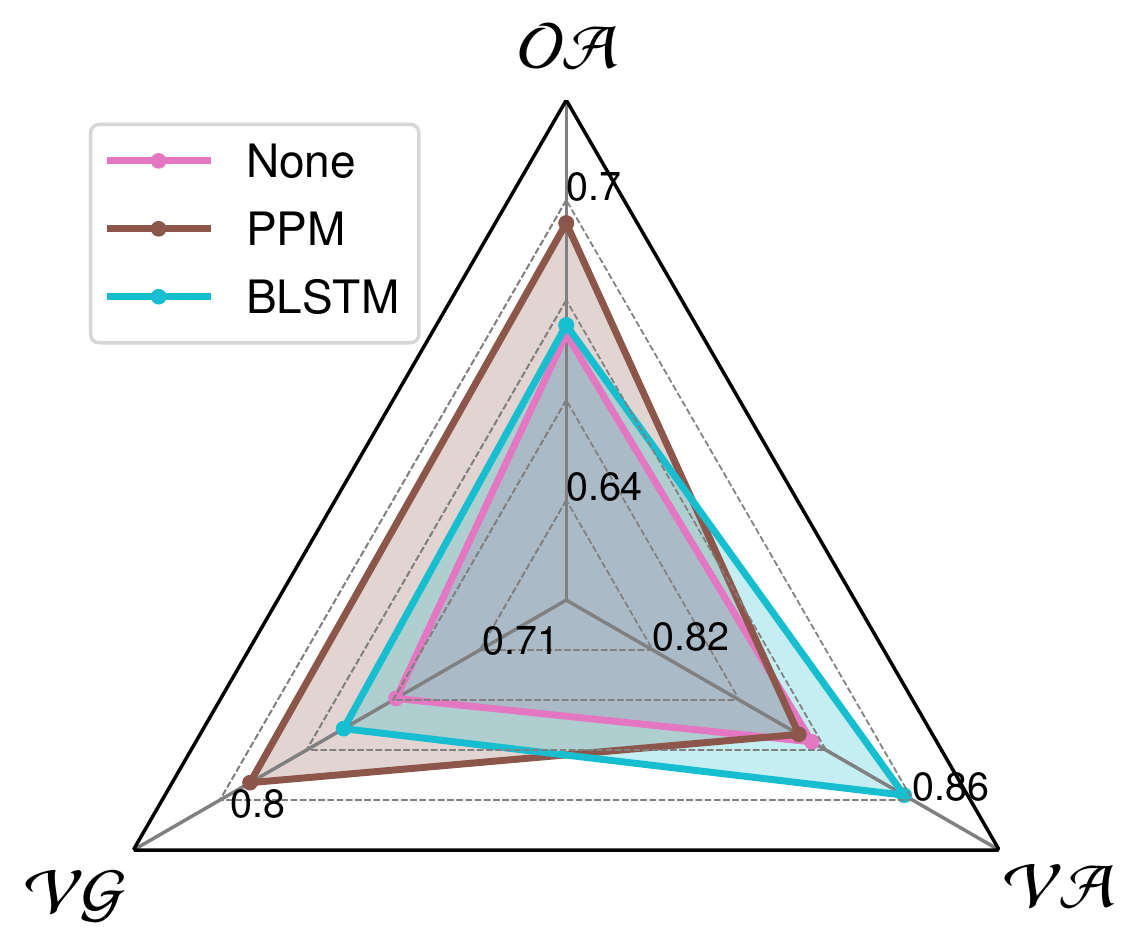}
        
        \caption{Attention-based PRED}\label{fig:Atten-pred}
   \end{subfigure}
   \vspace{-3mm}
   \caption{Comparison of CNTX modules.}
   \vspace{-4mm}
\end{figure}

Similar to PRED layers, the evaluation results of CNTX modules are illustrated in Fig.~\ref{fig:CTC-pred} and Fig.~\ref{fig:Atten-pred}. We find that the effect of CNTX modules in detail is highly coupled with prediction layers.

As stated in Sec.~\ref{comparison-prediction-layers}, attention-based decoders are more powerful in learning vocabulary from training data. Consequently, it brings less change in $\mathcal{VA}$ and $\mathcal{VG}$ to add more context modules to attention-based PRED layers. Besides, context modules, which perform as contextual information extractor, in fact, facilitates visual observation of attention-based and segmentation-based decoders.

As for CTC-family models, the situation is different. PPM and BLSTM significantly improve their $\mathcal{VA}$ and impair the $\mathcal{VG}$, as the CTC decoder itself lacks of proper context modeling. The performance change in the three metrics brought by context modules on CTC-family models is shown in Fig.~\ref{fig:CTC-pred}.

In summary, it is effective to strengthen the vocabulary learning of models with proper context modules: BLSTM for attention-based, PPM for CTC and segmentation-based decoder. After all, it is a trade-off between $\mathcal{VA}$ and $\mathcal{VG}$.

\subsection{Combination Recommendation}

Based on Tab.~\ref{tab:model-metrics} and the previous analyses, we recommend two combinations for different situations, depending on whether the vocabulary of target images are given.

Model \circlethree,  attention-based with BLSTM, achieves the best $\mathcal{VA}$ benefiting from the powerful CNTX module and PRED layer. This merit of model \circlethree in vocabulary learning also leads to the best $\mathcal{GA}$, corresponding to the performance on conventional benchmarks. It is evidenced by the high score in $\mathcal{VA}$ and $\mathcal{GA}$ that \circlethree can perform well in applications where the vocabulary of test images are mostly a restricted subset of training data. Accordingly, model \circlethree, similar to \cite{yang2019symmetry} in network design, is our first recommended combination for strong vocabulary learning ability.

As for many applications in the industry, algorithms trained with data in limited vocabulary are supposed to generalize well to more general words. Model \circleseven maintains good vocabulary generalization ability as it gets the best $\mathcal{VG}$. Therefore, we recommend the combination \circleseven, which is a CA-FCN-like~\cite{ca-fcn} structure, for scenarios where the generalization of vocabulary is concerned.


\section{Remedy by Mutual Learning}

Previous sections demonstrate the trade-off between $\mathcal{VA}$ and $\mathcal{VG}$ and the diverse advantages of models. In this section, we propose a simple yet effective training strategy for combining advantages of models in different prediction layers, i.e., attention-based and segmentation-based decoders.

The idea is basically inspired by knowledge distillation~\cite{hinton2015distilling} and deep mutual learning~\cite{zhang2018deep}. Similar to knowledge distillation, mutual learning of two models is a training strategy where models learn collaboratively. Knowledge distillation strategy transfers knowledge from a pre-trained powerful teacher network to student networks, while our approach optimizes two models simultaneously from scratch.

We choose the ensemble of the segmentation-based decoder and attention-based decoder as base models due to their advantages revealed in Fig.~\ref{fig:prediction-layers}. We suppose the generalization of segmentation-based decoders supervises attention-based decoders to learn to alleviate vocabulary reliance, and the accurate attention of attention-based decoders improves segmentation-based decoders in return.

\subsection{Optimization}

\begin{figure}[h]
    \centering
    \vspace{-3mm}
    \includegraphics[width=0.75\linewidth]{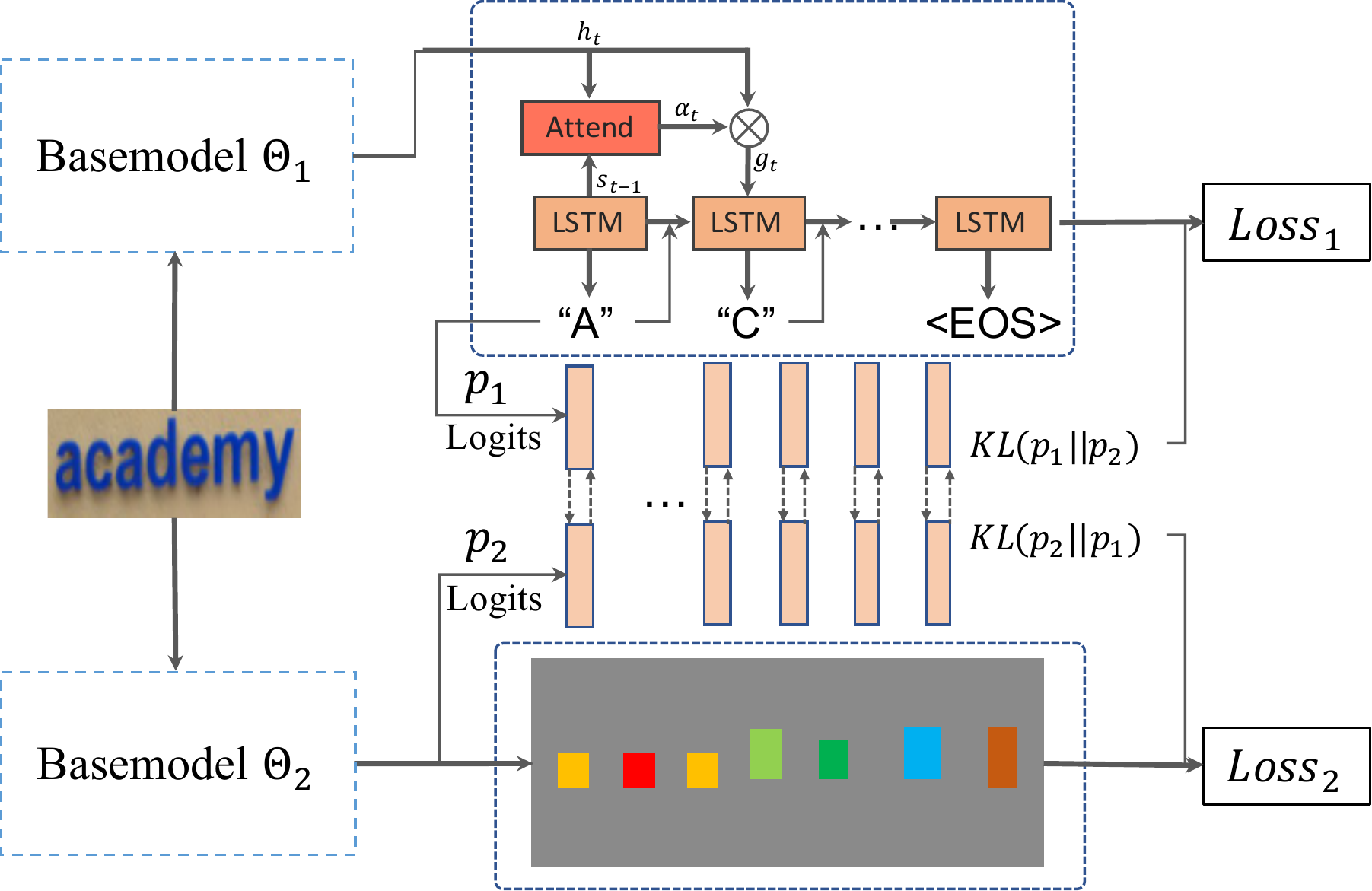}
    \vspace{-2mm}
    \caption{The mutual learning of attention-based decoder(top) and segmentation-based decoder(bottom). The KL divergence of logits are computed as auxiliary supervision, which makes the models learn collaboratively.}
    \vspace{-2mm}
    \label{fig:external}
\end{figure}

Let $\Theta_1$ and $\Theta_2$ be the network applying attention-based PRED layer and segmentation-based PRED layer, respectively. In addition to the original loss of the network $L_{\Theta_1}$ and $L_{\Theta_2}$, the Kullback Leibler (KL) Divergence is computed as an auxiliary loss. Then the ultimate loss for $\Theta_1$ and $\Theta_2$ are:
\vspace{-3mm}
\begin{equation}
    \begin{split}
        L_1 = \sum_{i}^{Y}{D_{KL}(p_2^i|| p_1^i)} + L_{\Theta_1} \\
        L_2 = \sum_{i}^{Y}{D_{KL}(p_1^i|| p_2^i)} + L_{\Theta_2}
    \end{split}
    \label{KL}
\end{equation}
where $p_1$, $p_2$ are the sequence of logits produced by $\Theta_1$ and $\Theta_2$, respectively. $D_{KL}$ is the KL Divergence and $Y$ is the sequential label. Note that for segmentation-based decoders, the logits are ``voted'' scores~\cite{ca-fcn} inside the shrunken region of characters.

From the Eq.~\ref{KL}, we can optimize the networks mutually supervised. The optimization is described in Alg.~\ref{alg:mutual}.

\vspace{-3mm}
\begin{algorithm}[ht]
\caption{Optimization of mutual learning.}

\SetAlgoLined
\SetKwInOut{input}{Input}\SetKwInOut{output}{Output}\SetKwInOut{data}{Data}\SetKwInOut{initialize}{Initialize}

\SetKwFunction{forward}{Forward}\SetKwFunction{backward}{Backward}

\input{Training data $X$ and label $Y$.}
\initialize{$\Theta_1$ and $\Theta_2$ separately.}

\While{not converged}{
    $p_1\leftarrow $ \forward{$\Theta_1$, $X$};
    
    $p_2\leftarrow $ \forward{$\Theta_2$, $X$};
    
    Compute $L_1$ from Eq.~\ref{KL};
    
    \backward{$\Theta_1$, $L_1$};
    
    $p_1\leftarrow $ \forward{$\Theta_1$, $X$};
    
    $p_2\leftarrow $ \forward{$\Theta_2$, $X$};
    
    Compute $L_2$ from Eq.~\ref{KL};
    
    \backward{$\Theta_2$, $L_2$};
}

\label{alg:mutual}
\end{algorithm}

\vspace{-0.1in}
\subsection{Experimental Validation}

\begin{table}[t]
    \vspace{-3mm}
    \caption{Performance comparison of mutual learning strategy on our metrics. ``Mut.'' indicates using mutual learning or not. The raw accuracy is shown in Tab.~\ref{tab:module-performance}.}
    \vspace{-6mm}
    \begin{center}
    \begin{tabular}{|c|c|c|ccc|c|}
    \hline
        No. & PRED &  Mut. &  
        $\mathcal{VA}$  &   $\mathcal{VG}$  &   $\mathcal{OA}$      &     $\mathcal{HM}$  \\
    \hline
        \circleone  &   Atten. & \xmark &
        83.2                & 77.1              & 69.6                  & 76.9           \\
        \circlenine  &   Atten. & \cmark &
        \textbf{85.8}                & \textbf{93.6}              & \textbf{71.5}                  & \textbf{82.6}           \\
        \hline
        \circleseven  &   Seg. & \xmark &
        75.8                & 97.3              & 69.9                  & 80.8           \\
        \circleten  &   Seg. & \cmark      &
        \textbf{82.3}                & 96.0              & \textbf{70.7}                  & \textbf{81.7}           \\
        
    \hline
    \end{tabular}
    \end{center}
    \vspace{-5mm}
    
    \vspace{-3mm}
    \label{tab:mutual-metrics}
\end{table}

We evaluate the mutual learning strategy using the proposed evaluation framework and exhibit the raw accuracy and performance on our metrics in Tab.~\ref{tab:module-performance} and Tab.~\ref{tab:mutual-metrics}, respectively. Experimental results demonstrate the significant improvement of base models brought by the mutual learning strategy.

These two models united by the mutual learning strategy maintains diverse properties and distinguishable advantage. The joint training procedure combines their inclination to visual features and vocabularies by harmonizing their estimation with the KL divergence. As evidence indicates, the $\mathcal{OA}$ and $\mathcal{VA}$ of both models are improved, which verifies the effectiveness of the mutual learning strategy.

Moreover, the vocabulary reliance of attention-based decoder is neutralized by the segmentation-based decoder. In the training of attention-based decoder, the prediction of the segmentation-based model, which relies more on visual features, acts as an extra visual regularization. In addition to minimizing $L_{\Theta_1}$, $\Theta_1$ is driven to fit the observation probability of $\Theta_2$. Quantitatively, the $\mathcal{VG}$ of $\Theta_1$ is boosted from 77.1\% to 93.6\%. In raw accuracy, the performance gap between images with words in and out of the vocabulary on LS data is almost halved (32.7\% to 16.9\%).

The qualitative comparison of the proposed mutual learning strategy is shown in Fig.~\ref{fig:prediction-mutual}. 
Notable improvement on benchmarks demonstrates the effectiveness of the proposed mutual learning strategy, thus validating it reasonable to integrate the advantages of different PRED layers.

\section{Conclusion}

In this paper, we investigate an important but long-neglected problem: vocabulary reliance in scene text recognition methods. A comprehensive framework is built for comparing and analyzing individual text recognition modules and their combinations. Based on this framework, a series of key observations and findings have been acquired, as well as valuable recommendations, which could be conducive to the future research of scene text recognition. Moreover, we have analyzed current contextual and prediction modules and proposed a mutual learning strategy for enhancing their vocabulary learning ability or generalization ability to words out of vocabulary.

\section*{Acknowledgement}
\vspace{-2mm}
This research was supported by National Key R\&D Program of China (No. 2017YFA0700800). 
The authors would like to thank Minghui Liao for helpful discussions.
{\small
\bibliographystyle{ieee_fullname}
\bibliography{egbib}

\begin{thebibliography}{10}\itemsep=-1pt

\bibitem{baek2019wrong}
Jeonghun Baek, Geewook Kim, Junyeop Lee, Sungrae Park, Dongyoon Han, Sangdoo
  Yun, Seong~Joon Oh, and Hwalsuk Lee.
\newblock What is wrong with scene text recognition model comparisons? dataset
  and model analysis.
\newblock {\em 2019 IEEE International Conference on Computer Vision}, 2019.

\bibitem{cheng2017fan}
Z. {Cheng}, F. {Bai}, Y. {Xu}, G. {Zheng}, S. {Pu}, and S. {Zhou}.
\newblock Focusing attention: Towards accurate text recognition in natural
  images.
\newblock In {\em {ICCV} 2017}, pages 5086--5094, Oct 2017.

\bibitem{aon}
Zhanzhan Cheng, Yangliu Xu, Fan Bai, Yi Niu, Shiliang Pu, and Shuigeng Zhou.
\newblock {AON:} towards arbitrarily-oriented text recognition.
\newblock In {\em {CVPR} 2018, Salt Lake City, UT, USA, June 18-22, 2018},
  pages 5571--5579, 2018.

\bibitem{nlpr_attn_ctc}
Yunze Gao, Yingying Chen, Jinqiao Wang, Hanqing Lu, 1National~Lab of
  Pattern~Recognition, Institute of Automation, Chinese~Academy of Sciences,
  Beijing, and China.
\newblock Reading scene text with attention convolutional sequence modeling.
\newblock 1709.

\bibitem{gao2019reading}
Yunze Gao, Yingying Chen, Jinqiao Wang, Ming Tang, and Hanqing Lu.
\newblock Reading scene text with fully convolutional sequence modeling.
\newblock {\em Neurocomputing}, 339:161--170, 2019.

\bibitem{ctc}
Alex Graves, Santiago Fern{\'a}ndez, Faustino Gomez, and J{\"u}rgen
  Schmidhuber.
\newblock Connectionist temporal classification: Labelling unsegmented sequence
  data with recurrent neural networks.
\newblock In {\em Proceedings of the 23rd International Conference on Machine
  learning}, pages 369--376, Pittsburgh, Pennsylvania, USA, 2006. IMLS.

\bibitem{synthtext}
Ankush Gupta, Andrea Vedaldi, and Andrew Zisserman.
\newblock Synthetic data for text localisation in natural images.
\newblock In {\em CVPR}, pages 2315--2324, 2016.

\bibitem{resnet}
Kaiming He, Xiangyu Zhang, Shaoqing Ren, and Jian Sun.
\newblock Deep residual learning for image recognition.
\newblock In {\em PAMI}, pages 770--778, 2016.

\bibitem{sliding_lstm_ctc2016}
Pan He, Weilin Huang, Yu Qiao, Chen~Change Loy, and Xiaoou Tang.
\newblock Reading scene text in deep convolutional sequences.
\newblock In {\em Proceedings of the Thirtieth {AAAI} Conference on Artificial
  Intelligence, February 12-17, 2016, Phoenix, Arizona, {USA.}}, pages
  3501--3508, 2016.

\bibitem{hinton2015distilling}
Geoffrey Hinton, Oriol Vinyals, and Jeff Dean.
\newblock Distilling the knowledge in a neural network.
\newblock {\em arXiv preprint arXiv:1503.02531}, 2015.

\bibitem{lstm}
Sepp Hochreiter and J{\"u}rgen Schmidhuber.
\newblock Long short-term memory.
\newblock {\em Neural computation}, 9(8):1735--1780, 1997.

\bibitem{hu2020gtc}
Wenyang Hu, Xiaocong Cai, Jun Hou, Shuai Yi, and Zhiping Lin.
\newblock Gtc: Guided training of ctc towards efficient and accurate scene text
  recognition.
\newblock {\em arXiv preprint arXiv:2002.01276}, 2020.

\bibitem{mjsynth}
Max Jaderberg, Karen Simonyan, Andrea Vedaldi, and Andrew Zisserman.
\newblock Synthetic data and artificial neural networks for natural scene text
  recognition.
\newblock {\em NIPS Deep Learning Workshop}, 2014.

\bibitem{karatzas2015icdar}
Dimosthenis Karatzas, Lluis Gomez-Bigorda, Anguelos Nicolaou, Suman Ghosh,
  Andrew Bagdanov, Masakazu Iwamura, Jiri Matas, Lukas Neumann,
  Vijay~Ramaseshan Chandrasekhar, Shijian Lu, et~al.
\newblock Icdar 2015 competition on robust reading.
\newblock In {\em 2015 13th ICDAR}, pages 1156--1160. IEEE, 2015.

\bibitem{ic13}
D. {Karatzas}, F. {Shafait}, S. {Uchida}, M. {Iwamura}, L.~G. i. {Bigorda},
  S.~R. {Mestre}, J. {Mas}, D.~F. {Mota}, J.~A. {Almazàn}, and L.~P. {de las
  Heras}.
\newblock Icdar 2013 robust reading competition.
\newblock In {\em 2013 12th ICDAR}, pages 1484--1493, Aug 2013.

\bibitem{Newsgroup20}
Tom~Mitchell Ken~Lang.
\newblock Newsgroup 20 dataset.
\newblock 1999.

\bibitem{khare2019novel}
Vijeta Khare, Palaiahnakote Shivakumara, Chee~Seng Chan, Tong Lu, Liang~Kim
  Meng, Hon~Hock Woon, and Michael Blumenstein.
\newblock A novel character segmentation-reconstruction approach for license
  plate recognition.
\newblock {\em Expert Systems with Applications}, 131:219--239, 2019.

\bibitem{krishnan2018word}
Praveen Krishnan, Kartik Dutta, and CV Jawahar.
\newblock Word spotting and recognition using deep embedding.
\newblock In {\em 2018 13th IAPR International Workshop on Document Analysis
  Systems (DAS)}, pages 1--6. IEEE, 2018.

\bibitem{lei2018scene}
Zhengchao Lei, Sanyuan Zhao, Hongmei Song, and Jianbing Shen.
\newblock Scene text recognition using residual convolutional recurrent neural
  network.
\newblock {\em Machine Vision and Applications}, 29(5):861--871, 2018.

\bibitem{li2018toward}
Hui Li, Peng Wang, and Chunhua Shen.
\newblock Toward end-to-end car license plate detection and recognition with
  deep neural networks.
\newblock {\em IEEE Transactions on Intelligent Transportation Systems},
  20(3):1126--1136, 2018.

\bibitem{Liao2019MaskTA}
Minghui Liao, Pengyuan Lyu, Minghang He, Cong Yao, Wenhao Wu, and Xiang Bai.
\newblock Mask textspotter: An end-to-end trainable neural network for spotting
  text with arbitrary shapes.
\newblock {\em IEEE transactions on pattern analysis and machine intelligence},
  2019.

\bibitem{Liao2019RealtimeST}
Minghui Liao, Zhaoyi Wan, Cong Yao, Kai Chen, and Xiang Bai.
\newblock Real-time scene text detection with differentiable binarization.
\newblock {\em ArXiv}, abs/1911.08947, 2019.

\bibitem{ca-fcn}
Minghui {Liao}, Jian {Zhang}, Zhaoyi {Wan}, Fengming {Xie}, Jiajun {Liang},
  Pengyuan {Lyu}, Cong {Yao}, and Xiang {Bai}.
\newblock Scene text recognition from two-dimensional perspective.
\newblock In {\em {AAAI}}, 2019.

\bibitem{long2018scene}
Shangbang Long, Xin He, and Cong Yao.
\newblock Scene text detection and recognition: The deep learning era.
\newblock {\em arXiv preprint arXiv:1811.04256}, 2018.

\bibitem{Long2018TextSnakeAF}
Shangbang Long, Jiaqiang Ruan, Wenjie Zhang, Xin He, Wenhao Wu, and Cong Yao.
\newblock Textsnake: A flexible representation for detecting text of arbitrary
  shapes.
\newblock In {\em ECCV}, 2018.

\bibitem{mask_textspotter}
Pengyuan Lyu, Minghui Liao, Cong Yao, Wenhao Wu, and Xiang Bai.
\newblock Mask textspotter: An end-to-end trainable neural network for spotting
  text with arbitrary shapes.
\newblock In {\em ECCV}, pages 67--83, 2018.

\bibitem{mishra2012scene}
Anand Mishra, Karteek Alahari, and CV Jawahar.
\newblock Scene text recognition using higher order language priors.
\newblock In {\em BMVC-British Machine Vision Conference}. BMVA, 2012.

\bibitem{quy2013recognizing}
T.~Q. {Phan}, P. {Shivakumara}, S. {Tian}, and C.~L. {Tan}.
\newblock Recognizing text with perspective distortion in natural scenes.
\newblock In {\em 2013 IEEE International Conference on Computer Vision}, pages
  569--576, Dec 2013.

\bibitem{ren2016cnn}
Xiaohang Ren, Kai Chen, and Jun Sun.
\newblock A cnn based scene chinese text recognition algorithm with synthetic
  data engine.
\newblock {\em arXiv preprint arXiv:1604.01891}, 2016.

\bibitem{cute}
Anhar Risnumawan, Palaiahankote Shivakumara, Chee~Seng Chan, and Chew~Lim Tan.
\newblock A robust arbitrary text detection system for natural scene images.
\newblock {\em Expert Systems with Applications}, 41(18):8027 -- 8048, 2014.

\bibitem{shi2017end}
Baoguang Shi, Xiang Bai, and Cong Yao.
\newblock An end-to-end trainable neural network for image-based sequence
  recognition and its application to scene text recognition.
\newblock {\em PAMI}, 39(11):2298--2304, 2017.

\bibitem{Shi2016RobustST}
Baoguang Shi, Xinggang Wang, Pengyuan Lyu, Cong Yao, and Xiang Bai.
\newblock Robust scene text recognition with automatic rectification.
\newblock {\em 2016 IEEE Conference on Computer Vision and Pattern Recognition
  (CVPR)}, pages 4168--4176, 2016.

\bibitem{aster}
Baoguang Shi, Mingkun Yang, Xinggang Wang, Pengyuan Lyu, Cong Yao, and Xiang
  Bai.
\newblock Aster: An and attentional scene and text recognizer and with flexible
  and rectification.
\newblock In {\em PAMI}, pages 1--1. IEEE, 2018.

\bibitem{su2014accurate}
Bolan Su and Shijian Lu.
\newblock Accurate scene text recognition based on recurrent neural network.
\newblock In Daniel Cremers, Ian Reid, Hideo Saito, and Ming-Hsuan Yang,
  editors, {\em Computer Vision -- ACCV 2014}, pages 35--48, Cham, 2015.
  Springer International Publishing.

\bibitem{tian2017unified}
Shu Tian, Xu-Cheng Yin, Ya Su, and Hong-Wei Hao.
\newblock A unified framework for tracking based text detection and recognition
  from web videos.
\newblock {\em IEEE transactions on pattern analysis and machine intelligence},
  40(3):542--554, 2017.

\bibitem{Wan2019TextScannerRC}
Zhaoyi Wan, Mingling He, Haoran Chen, Xiang Bai, and Cong Yao.
\newblock Textscanner: Reading characters in order for robust scene text
  recognition.
\newblock {\em ArXiv}, abs/1912.12422, 2019.

\bibitem{wang2011end}
Kai Wang, Boris Babenko, and Serge Belongie.
\newblock End-to-end scene text recognition.
\newblock In {\em ICCV}, ICCV '11, pages 1457--1464, Washington, DC, USA, Nov
  2011. IEEE Computer Society.

\bibitem{xie2016fully}
Zecheng Xie, Zenghui Sun, Lianwen Jin, Ziyong Feng, and Shuye Zhang.
\newblock Fully convolutional recurrent network for handwritten chinese text
  recognition.
\newblock In {\em 2016 23rd International Conference on Pattern Recognition
  (ICPR)}, pages 4011--4016. IEEE, 2016.

\bibitem{Yang2019SymmetryConstrainedRN}
Mingkun Yang, Yushuo Guan, Minghui Liao, Xin He, Kaigui Bian, Song Bai, Cong
  Yao, and Xiang Bai.
\newblock Symmetry-constrained rectification network for scene text
  recognition.
\newblock {\em 2019 IEEE/CVF International Conference on Computer Vision
  (ICCV)}, pages 9146--9155, 2019.

\bibitem{yang2019symmetry}
MingKun Yang, Yushuo Guan, Minghui Liao, Xin He, Kaigui Bian, Song Bai, Cong
  Yao, and Xiang Bai.
\newblock Symmetry-constrained rectification network for scene text
  recognition.
\newblock {\em 2019 IEEE International Conference on Computer Vision}, 2019.

\bibitem{fcn_attn}
Xiao Yang, Dafang He, Zihan Zhou, Daniel Kifer, and C.~Lee Giles.
\newblock Learning to read irregular text with attention mechanisms.
\newblock In {\em Proceedings of the 26th International Joint Conference on
  Artificial Intelligence}, IJCAI'17, pages 3280--3286. AAAI Press, 2017.

\bibitem{Yao2014AUF}
Cong Yao, Xiang Bai, and Wenyu Liu.
\newblock A unified framework for multi-oriented text detection and
  recognition.
\newblock 2014.

\bibitem{Yao2012DetectingTO}
Cong Yao, Xiang Bai, Wenyu Liu, Yi Ma, and Zhuowen Tu.
\newblock Detecting texts of arbitrary orientations in natural images.
\newblock {\em 2012 IEEE Conference on Computer Vision and Pattern
  Recognition}, pages 1083--1090, 2012.

\bibitem{Yao2014StrokeletsAL}
Cong Yao, Xiang Bai, Baoguang Shi, and Wenyu Liu.
\newblock Strokelets: A learned multi-scale representation for scene text
  recognition.
\newblock {\em 2014 IEEE Conference on Computer Vision and Pattern
  Recognition}, pages 4042--4049, 2014.

\bibitem{ye2014text}
Qixiang Ye and David Doermann.
\newblock Text detection and recognition in imagery: A survey.
\newblock {\em IEEE transactions on pattern analysis and machine intelligence},
  37(7):1480--1500, 2014.

\bibitem{zhan2018verisimilar}
Fangneng Zhan, Shijian Lu, and Chuhui Xue.
\newblock Verisimilar image synthesis for accurate detection and recognition of
  texts in scenes.
\newblock In {\em Proceedings of the European Conference on Computer Vision
  (ECCV)}, pages 249--266, 2018.

\bibitem{zhang2019sequence}
Yaping Zhang, Shuai Nie, Wenju Liu, Xing Xu, Dongxiang Zhang, and Heng~Tao
  Shen.
\newblock Sequence-to-sequence domain adaptation network for robust text image
  recognition.
\newblock In {\em Proceedings of the IEEE Conference on Computer Vision and
  Pattern Recognition}, pages 2740--2749, 2019.

\bibitem{zhang2018deep}
Ying Zhang, Tao Xiang, Timothy~M Hospedales, and Huchuan Lu.
\newblock Deep mutual learning.
\newblock In {\em Proceedings of the IEEE Conference on Computer Vision and
  Pattern Recognition}, pages 4320--4328, 2018.

\bibitem{pspnet}
H. {Zhao}, J. {Shi}, X. {Qi}, X. {Wang}, and J. {Jia}.
\newblock Pyramid scene parsing network.
\newblock In {\em 2017 IEEE Conference on Computer Vision and Pattern
  Recognition (CVPR)}, pages 6230--6239, Honolulu, HI, USA, July 2017.

\bibitem{Zhou2017EASTAE}
Xinyu Zhou, Cong Yao, He Wen, Yuzhi Wang, Shuchang Zhou, Weiran He, and Jiajun
  Liang.
\newblock East: An efficient and accurate scene text detector.
\newblock {\em 2017 IEEE Conference on Computer Vision and Pattern Recognition
  (CVPR)}, pages 2642--2651, 2017.

\end{thebibliography}
}

\end{document}